\documentclass[runningheads]{llncs}
\usepackage{amsmath,amsfonts}
\usepackage{array}
\usepackage{ulem}
\usepackage[caption=false,font=normalsize,labelfont=sf,textfont=sf]{subfig}
\usepackage{textcomp}
\usepackage{stfloats}
\usepackage{url}
\usepackage{verbatim}
\usepackage{mathrsfs}
\usepackage[section]{placeins}
\usepackage{graphicx}
\usepackage{cite}
\usepackage{graphicx}
\usepackage{amsmath,amsfonts}	
\usepackage{xspace}
\usepackage{color}    
\usepackage{verbatim} 
	\usepackage{booktabs}  
	\usepackage{multirow}  
	\usepackage{graphicx}
	\usepackage{subfloat}

	\usepackage{diagbox}
	\usepackage{xspace}
	\usepackage{amssymb}
	\usepackage{multirow}
	\usepackage{weiwAlgorithm}
	\usepackage{color} 

\begin{document}

\title{Mlinear: Rethink the Linear Model for Time-series Forecasting}

\author{Wei Li\inst{1}, Xiangxu Meng\inst{1}, Chuhao Chen\inst{1} \and Jianing Chen\inst{1}\thanks{Corresponding author}
}

\institute{College of Computer Science and Technology, Harbin Engineering University, Harbin 150001, China\\
\email{\{wei.li, mxx, chenchuhao, kaqiz\}@hrbeu.edu.cn}}

\maketitle              
%

\begin{abstract}
	Recently, significant advancements have been made in time-series forecasting research, with an increasing focus on analyzing the nature of time-series data, e.g, channel-independence (CI) and channel-dependence (CD), rather than solely focusing on designing sophisticated forecasting models.
	However, current research has primarily focused on either CI or CD in isolation, and the challenge of effectively combining these two opposing properties to achieve a synergistic effect remains an unresolved issue.
	In this paper, we carefully examine the opposing properties of CI and CD, and raise a practical question that has not been effectively answered, e.g., ``How to effectively mix the CI and CD properties of time series to achieve better predictive performance?'' To answer this question, we propose Mlinear (MIX-Linear), a simple yet effective method based mainly on linear layers.
	The design philosophy of Mlinear mainly includes two aspects: (1) dynamically tuning the CI and CD properties based on the time semantics of different input time series, and (2) providing deep supervision to adjust the individual performance of the ``CI predictor'' and ``CD predictor''. In addition, empirically, we introduce a new loss function that significantly outperforms the widely used mean squared error (MSE) on multiple datasets.
	Experiments on time-series datasets covering multiple fields and widely used have demonstrated the superiority of our method over PatchTST which is the lateset Transformer-based method in terms of the MSE and MAE metrics on 7 datasets with identical sequence inputs (336 or 512). Specifically, our method significantly outperforms PatchTST with a ratio of 21:3 at 336 sequence length input and 29:10 at 512 sequence length input. Additionally, our approach has a 10 $\times$ efficiency advantage at the unit level, taking into account both training and inference times. 

\end{abstract}

\section{Introduction}
Analyzing past data to predict future trends is a classic research field with significant application value in both industry and academia. As a hot topic in time series forecasting, multivariate forecasting has received great attention and made remarkable progress in recent years.
Over the past few decades, solutions for time-series forecasting (TSF) have undergone a transition from traditional statistical methods such as ARIMA~\cite{ariyo2014stock}, and machine learning techniques such as GBRT~\cite{friedman2001greedy}, towards deep learning-based approaches. 
In addition, due to the knowledge capacity advantage of DL-based models, sequence prediction has gradually expanded to broader fields, e.g., long sequence time-series forecasting (LSTF).
In particular, novel techniques such as Recurrent Neural Networks~\cite{hochreiter1997long}, Temporal Convolutional Networks~\cite{bai2018empirical} and Transformer~\cite{vaswani2017attention} have been widely applied in this field.
Among various brilliant LSTF methods, the Transformer is undoubtedly the most remarkable one.
Stemming from natural language processing, Transformer-based models~\cite{vaswani2017attention} have achieved strong results in computer vision~\cite{dosovitskiy2020image,lin2022swintrack,liu2022swin,guo2022attention,lin2022survey,liang2021swinir}, graph data processing~\cite{yun2019graph,cai2020graph,min2022transformer}, and various downstream tasks~\cite{xu2021improved,li2022transformer,lei2021transformer,valanarasu2021medical,shamshad2023transformers}. 
The Transformer's exceptional ability to perform global modeling allows it to focus on long-span relationships between points in a long sequence, making it capable of capturing long-term temporal dependencies.
%
%
%
These Transformer-based LSTF approachs have made significant progress, even achieving practical applications during the Beijing Winter Olympics~\cite{wu2021autoformer}.
To our knowledge, two of the most notable methods in these type approachs are Informer and Autoformer. 
This can also be supported from the perspective of citation count.
Informer~\cite{zhou2021informer} tackles the issue of long-tail distribution of active attention in processing sequential data by proposing a sparse attention calculation method based on this property. This reduces the time complexity from $O(L^2)$ to $O(LlogL)$. On the other hand, Autoformer uses sequence factorization as a preprocessing method and replaces the dot-product attention mechanism with the Auto-Correlation Mechanism, achieving sequence-level connectivity and reducing the time complexity to $O(LlogL)$.
At the same time, there are also some other excellent Transformer-based LSTF methods that attempt to achieve accurate long-term sequence prediction from different perspectives, such as frequency domain attention and semantic attention. 
Fedformer~\cite{zhou2022fedformer}, Pyraformer~\cite{liu2021pyraformer}, and Triformer~\cite{cirstea2022triformer} are among the representative works in this regard.

%

%
Although Transformer-based methods have achieved significant predictive performance, it is important to note that most research has mainly followed the standard encoder-decoder architecture. 
While various works claim to have made improvements from semantic and frequency domain perspectives, a closer examination of past work reveals that the improvements mainly focused on efficiency, particularly in the calculation of attention mechanisms. 
In other words, these impressive results were mainly obtained through the global modeling capability of the multi-head attention mechanism, without a deep analysis of the essence of time series to achieve an improvement in forecasting accuracy.
A significant difference can be observed in recent research, where the introduction of channel-independent (CI) and channel-dependent (CD) ideas in time-series forecasting has led to some works achieving extremely remarkable performance using simple linear combinations~\cite{zeng2022transformers} and canonical Transformers~\cite{nie2022time}.
%
%
For example, PatchTST~\cite{nie2022time} introduces the concept of CI into the Transformer structure and combines it with the idea of patch to model local time dependencies. PatchTST does not make major improvements to the Transformer structure, but achieves performance beyond that of previous well-designed Transformer works using only a canonical Transformer encoder.
Compared to PatchTST, the performance of Dlinear~\cite{zeng2022transformers} may be even more surprising. With the concept of CI, they achieve significant advancement over Transformer-based LSTF methods with extremely low computational cost by simply stacking linear layers.

%
Take a deep understanding of the concepts of CI and CD for multivariate long time series prediction. 
A time series $L$ $\in\mathbb{R}^{L\times N}$ with $L$ time units and $N$ variables can be considered as a collection of individual channels, as shown in Figure B. As each variable comes from a different performance metric of a specific prediction task, there exist hidden and complex dependencies among them. CD means modeling the dependencies between different variables by treating the entire sequence as a whole input. In contrast, CI means treating each variable as an individual sequence and focusing on modeling the internal dependencies within each variable.

Although recent works have observed the significant advantage of CI over CD in LSTF tasks and abandoned CD to fully incorporate the properties of CI, we believe that this approach is not feasible as it completely disregards the modeling relationship between different variables through CD, leading to information bottleneck issues. 
Furthermore, we argue that the significant advantage of CI is only a product of specific structures.
As shown in Table~\ref{TABLE1}, for our proposed MLinear, there is no significant advantage difference between CD and CI, and each has its own advantages on different datasets. Therefore, in this paper, we focus on the relationship between CI and CD and raise a challenging but practical problem: \textit{\textbf{How to effectively mix CI and CD to achieve joint advantages?}} As shown in Table~\ref{TABLE1}, we present a simple mixing scheme and the final MLinear, which significantly improves performance.

In this paper, we aim to address this question by proposing a mixed CI and CD time series linear, MLinear model, that incorporates three essential design components:
\begin{table*}[h]
	\centering
	\begin{tabular}{cc|c|c|c|c}
		\toprule
		\multicolumn{2}{c|}{Methods}&\multicolumn{1}{c|}{CD} & \multicolumn{1}{c|}{CI} & \multicolumn{1}{c|}{MIX} &\multicolumn{1}{c}{MLinear}\\ \midrule
		\multicolumn{1}{c|}{\multirow{4}{*}{\rotatebox{90}{Electricity}}}&96       &0.141	&	0.135&	0.134&	\textbf{0.133}
		\\ 
		\multicolumn{1}{c|}{}&192 &0.154	&0.150&	\textbf{0.149}&	\textbf{0.149}
		\\ 
		\multicolumn{1}{c|}{}&336  &0.170	&	0.167&	0.166&	\textbf{0.165}
		\\ 
		\multicolumn{1}{c|}{}&720  &0.209&	0.206&	0.205	&\textbf{0.203}
		
		\\ \midrule
		
		\multicolumn{1}{c|}{\multirow{4}{*}{\rotatebox{90}{ETTh1}}}&96    & 0.364		&0.372		&0.367&	\textbf{0.359}
		
		\\ 
		\multicolumn{1}{c|}{}&192   &0.406	&0.413&	0.407&	\textbf{0.400}
		\\ 
		\multicolumn{1}{c|}{}&336   &0.429		&0.433&	0.428&	\textbf{0.417}
		\\ 
		\multicolumn{1}{c|}{}&720   & 0.425&	0.440&	0.434&	\textbf{0.429}
		
		\\ \midrule
		\multicolumn{1}{c|}{\multirow{4}{*}{\rotatebox{90}{ETTm1}}}&96   & 0.297		&0.292	&	0.291&	\textbf{0.289}
		
		\\ 
		\multicolumn{1}{c|}{}&192  & 0.335	&0.335&	0.334	&	\textbf{0.331}
		\\ 
		\multicolumn{1}{c|}{}&336   & \textbf{0.369}	&	0.373	&	0.372&	\textbf{0.369}	
		\\ 
		\multicolumn{1}{c|}{}&720 &\textbf{0.427}	&	0.431&	0.430&	0.428
		\\ \midrule
		\multicolumn{2}{c|}{Count}&\multicolumn{1}{c|}{3}&\multicolumn{1}{c|}{3}&\multicolumn{1}{c|}{10}&\multicolumn{1}{c}{18}\\
		\bottomrule
		
	\end{tabular}
	\caption{A brief comparison of CI, CD, linear aggregation, and Mlinear on three datasets is presented. The loss used is our proposed loss.}
	\label{TABLE1}
\end{table*}
%
\subsubsection{Mixing multi-channel and single-channel time series data}
Traditionally, when performing time-series forecasting, multivariate (multi-channel) time series data are considered as a whole input, which has become a conventional approach adopted by most Transformer-based models.
Due to the interdependence among multiple variables, this approach is commonly referred to as channel-dependence (CD) in recent works~\cite{nie2022time,zeng2022transformers}, as it manifests as the interdependence among different channels in deep learning models.
However, CD may lead to data pollution and distribution drift due to different data distributions among different channels. 
Dlinear breaks this approach and introduces channel independent (CI) forecasting of multi-channel variate. By considering each time series data channel separately, this approach subverts the input method and achieves significant performance improvement. 
However, despite the promising predictive performance of CI, it has not been thoroughly explored. Furthermore, the physical meaning of CD should not be completely overlooked. 
\subsubsection{Dynamically tuning CI and CD based on the temporal semantics of different sequences}
A natural idea is to mix CI and CD to do collaborative forecasting, but naively mixing them using a fixed set of parameters can lead to insufficient generalization, as the time dependencies among different input sequences vary.
For example, the temporal semantics of different time series inputs vary, such as in the case of peak points, which may mark the end or beginning of a period, and subsequences near a peak point typically have sustained high values over a period of time, while peak points near a trough point exhibit the opposite behavior. 
To address this issue, we propose a mix approach that can dynamically model CI and CD based on the temporal semantics of different inputs while maintaining extremely low parameter usage and computational complexity.
The performance differences of mixing using different methods are shown in Table~\ref{TABLE1}. 
%
\subsubsection{Deep supervision for CI and CD}
Mixing CI and CD intuitively means combining the predictions generated by the two components to achieve better forecasting performance. By obtaining two better individual performances, we can achieve a better joint performance. Deep supervision can effectively supervise the parameters at deeper levels. Therefore, the non-shared deep parameters of CI and CD can be better trained, resulting in better individual predictors for CI and CD and subsequently better joint performance. To our knowledge, this is the first attempt in the LSTF field to improve fitting performance through deep supervision.
\subsubsection{More powerful loss function}
In time series forecasting methods, a conventional approach is to use the mean squared error (MSE) loss for different models, as seen in Informer, Autoformer, Fedformer, PatchTST, and Dlinear. However, our experiments show that MSE loss performs poorly in the field of time series forecasting. A large body of research demonstrates that MSE loss tends to punish larger points, giving them more weight and ignoring the role of smaller points. If there are outliers in the sample, MSE loss will assign higher weights to them, but this is achieved at the expense of sacrificing the forecasting performance of other normal data points, leading to a decrease in the overall performance of the trained model. Based on this observation, we introduce a new loss function that significantly improves the robustness and forecasting performance of the model compared to MSE and MAE loss.

%
We introduce the MLinear model in greater detail and provide comprehensive experimental results in the following sections to conclusively support our claims. We demonstrate the effectiveness of our model through forecasting results and ablation studies.

\section{Related work}
\begin{figure*}[htbp]
	\hspace{5cm}
	\centering
	\includegraphics[width=1\textwidth]{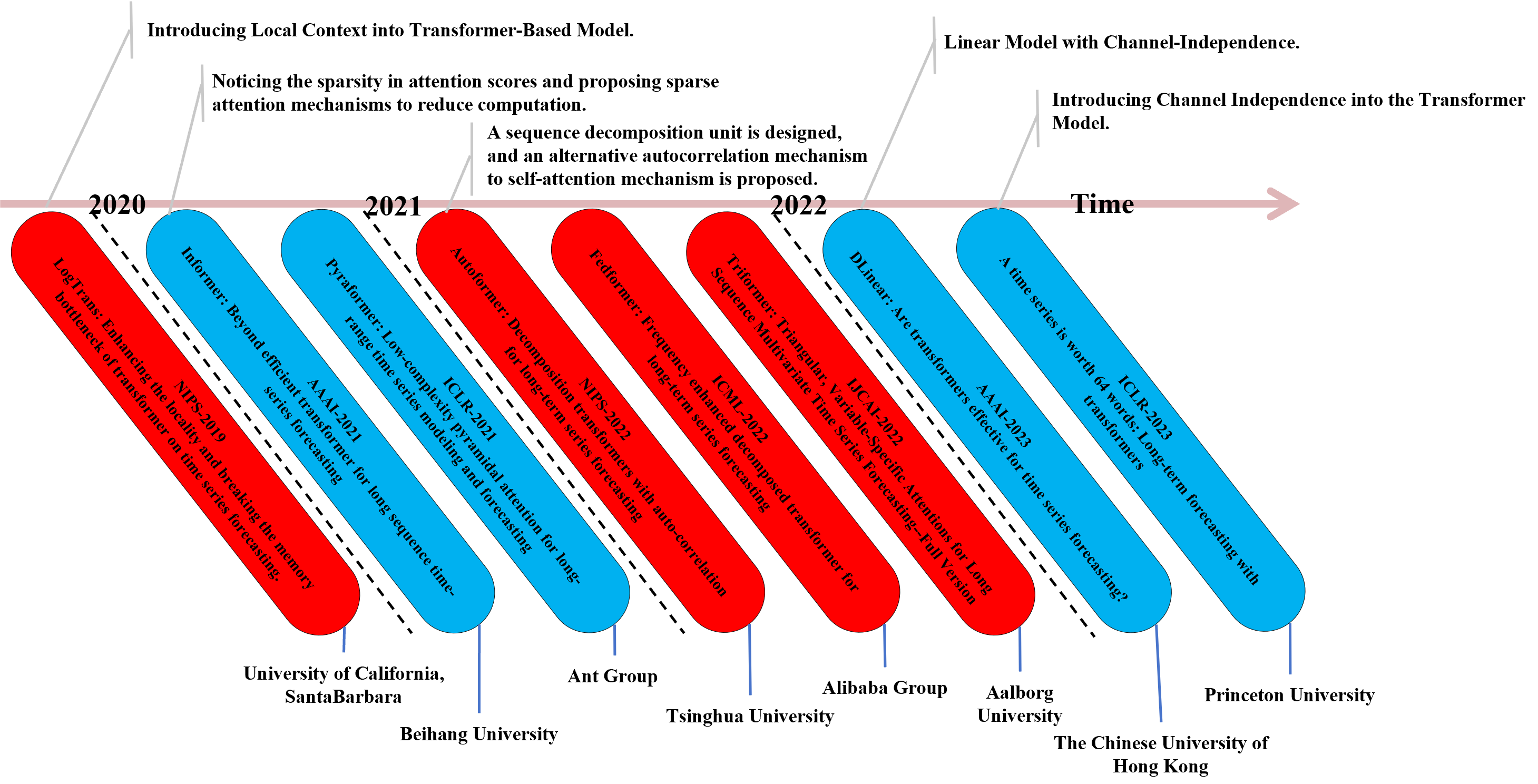}
	\caption{Milestone timeline for time series forecasting.}
	\label{Milestone}
\end{figure*}

In recent years, learning-based LSTF methods have made remarkable progress. From a model perspective, Transformer-based methods have emerged like mushrooms, with various excellent attention mechanisms and improved architectures flooding the field and dominating for a long time. However, with the attention of researchers to the CI and CD properties, non-Transformer-based methods have gained further development potential. Therefore, we will discuss the progress of relevant LSTF methods from these two perspectives.
In addition, the timeline of some milestone works is shown in Figure~\ref{Milestone}.
\subsection{Transformer-based Long-term Time Series Forecasting.}
In recent years, Transformer-based long sequence time-series forecasting (LSTF) methods have gained increasing attention due to their highly acclaimed ``Global modeling'' capability. The self-attention mechanism allows the model to automatically capture dependencies between different time steps in the time series data, enabling the Transformer model to better understand the inherent temporal dependencies in time series data.

Although Informer~\cite{zhou2021informer} is not the earliest, it is still considered the most notable one among them.
What makes Informer particularly impressive is its exploration of the long-tail distribution problem in the global attention mechanism, which led to the proposal of a sparse attention mechanism. 
Sparse-self-attention analyzes the sparsity of temporal dependencies in time series data and employs a small number of sampled computations to obtain query matrices with significant attention scores. 
By using a limited number of active queries for $query\times key$ multiplication, the time complexity of the self-attention mechanism is reduced from $O(L^{2})$ to $O(LlogL)$.
Autoformer~\cite{wu2021autoformer} introduced a decomposition architecture that can extract more predictable components from complex temporal patterns. Based on stochastic process theory, Auto-Correlation Mechanism is proposed to replace the point-to-point attention mechanism, achieving series-wise connectivity and a time complexity of $O(LlogL)$, breaking the bottleneck of information utilization in sparse attention mechanisms.
Fedformer~\cite{zhou2022fedformer} utilizes the fact that most time series tend to have sparse representation in well-known bases (such as Fourier transform) and proposes an Enhance-frequency Transformer to improve the performance of long-term time series prediction. 
Pyraformer~\cite{liu2021pyraformer} applies a pyramidal attention module with inter-scale and intra-scale connections, achieving linear complexity.
The latest Transformer-based LSTF work, PatchTST~\cite{nie2022time}, combines the channel-independent property of time series with the idea of patches, focusing on a temporal subset for modeling and achieving significant prediction performance. 
The aforementioned works are all in the general domain, and in addition, the Transformer has also demonstrated its modeling ability for long sequences in specific fields, such as stock trend forecasting~\cite{ni2021hybrid,ding2020hierarchical,zhang2022transformer} and traffic flow forecasting~\cite{cai2020traffic,yan2021learning,xu2020spatial}.
Overall, these models have demonstrated the effectiveness of the Transformer architecture in tackling the challenges of time series forecasting, and have opened up exciting new possibilities for future research in this field.

\subsection{Non-Transformer-based Long-term Time Series Forecasting.}
As far as we know, the earliest work focused on time series prediction can be traced back to ARIMA~\cite{ariyo2014stock}, which follows a Markov process and establishes an autoregressive model for recursive sequence prediction. The model is very simple, requiring only endogenous variables without the need for other exogenous variables.
However, autoregressive processes are not sufficient to handle nonlinear and non-stationary sequences. Moreover, they require time series data to be stationary or stationary after differencing. Therefore, ARIMA can only capture linear relationships and cannot capture nonlinear relationships. 

With the rapid development of neural networks and their extraordinary capabilities in other fields, RNN-based methods~\cite{petnehazi2019recurrent,hewamalage2021recurrent,khaldi2023best,dudukcu2023temporal} have also emerged, thanks to their ability to capture the sequential dependencies between input and output sequences, flexibility in handling variable-length sequences, and modeling of complex patterns. 
For example, GCAR~\cite{geng2022graph} is a model designed for multivariate time series prediction, and its innovation lies in enhancing the memory of key features and capturing time-varying patterns through reliable interaction. This model utilizes feature-level and multi-level attention mechanisms to extract multi-head temporal correlations and distinguish the contributions of external features. Additionally, GCAR employs two parallel LSTMs to capture continuous dynamic changes in different series and uses a fusion gate to balance information conflicts.
%
%
Unlike traditional RNN models, the Temporal Convolutional Network (TCN)~\cite{liu2021time,hewage2020temporal,fan2021parallel,wang2020short} does not need to explicitly maintain historical state information, but instead captures local and global dependencies of time series data through convolutional layers.
For instance, SCINet~\cite{liu2021time} is a TCN-based method for time series prediction that uses sample convolution and interaction for temporal modeling. It expands the receptive field of the convolution operation and enables multi-resolution analysis through a downsample-convolve-interact architecture, improving the extraction of temporal relation features and prediction accuracy.

The above methods attempt to provide a non-Transformer alternative for long-term time series prediction from statistical, RNN, and TCN perspectives. However, when considering a comprehensive range of fields and longer sequence lengths, Transformer-based LSTF methods still dominate in most cases. 
Nevertheless, the emergence of DLinear~\cite{zeng2022transformers} has shaken the Transformer family, as its simple linear layer combined with channel independence properties has allowed it to outperform almost all Transformer-based LSTF methods with extremely low computational cost. This suggests that non-Transformer based methods are worth exploring further.
\section{Methodology}
\label{sec:approaches}
\begin{figure*}[htbp]
	\hspace{4cm}
	\centering
	\includegraphics[width=1\textwidth]{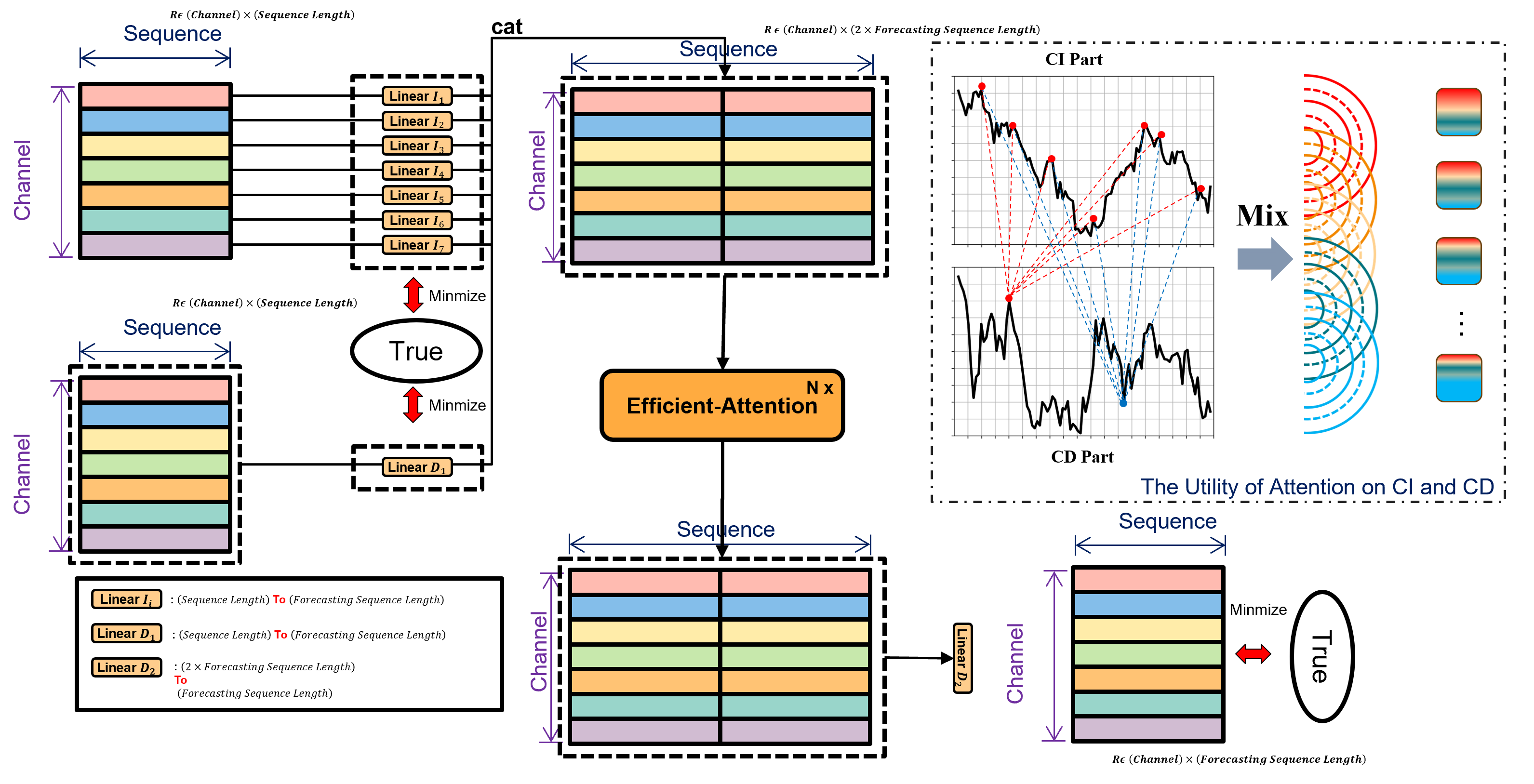}
	\caption{Model architecture of MLinear.}
	\label{fig:example}
\end{figure*}
\subsection{The Physical Meaning of CI and CD}
CD and CI both perform sequence modeling through a linear layer, and the linear layers used in both architectures have the same composition, with the input size equal to the sequence length and the output size equal to the forecasting-sequence length. 
The difference between them lies in the input they take. 
The input $x_i$ to every CI linear is of shape $\in\mathbb{R}^{1\times L}$, and $i=n$, while the input $X$ to CD linear is of shape $\in\mathbb{R}^{n\times L}$. 
In CD, the weights of the linear layer are shared across the $n$ sequences, meaning that the weights model the dependencies between different channels. 
In CI, there are $n$ linear layers, each modeling a single-channel sequence meaning that weight learning occurs only on a single-channel sequence.
Considering the input $X=[x_1, x_2, ..., x_n]$, the operations and concat processes of CI and CD are shown as follows:
\begin{equation}
	\begin{aligned}
	X_{\text{CI}} &= \operatorname{Concat}(x_1W_{{\text{CI}}_1},...,x_iW_{{\text{CI}}_i}),W_{{\text{CI}}_i}\in \mathbb{R}^{L\times S}\\	
	X_{\text{CD}} &= XW_{{\text{CD}}}, W_{\text{CD}}\in \mathbb{R}^{L\times S}\\
	X_{\text{CI+CD}} &= \operatorname{Concat}(X_{\text{CI}},X_{\text{CD}})
	\end{aligned}
\end{equation}
where $W_i$ represents the linear weights corresponding to CI and CD. 
\subsection{Time-Semantics-Aware Channel Mixing Modulate}
To avoid treating all input time series equally by using fixed weights, it is necessary to dynamically adjust the relationship between the temporal semantics provided by CI and CD in order to aggregate forecasting-beneficial temporal dependencies more appropriately.
We firstly discuss the differences between fixed weights and time-semantics-aware (TSA) weights, and then demonstrate the process of our proposed method to tune different temporal semantics.
\subsubsection{Fixed or TSA-weights  modulate.}
Most previous work on temporal modeling has been based on Transformers or their variants. 
While the focus of these different works varies, they all share the common idea of dynamically modeling dependencies on different input sequences. 
This idea of dynamic modeling is not only applicable to time-series forecasting tasks, but also to various attention mechanisms in other communities.
The latest temporal LSTF model, DLinear, achieves excellent performance by using a simple linear layer with fixed weights combined with the property of CI. 
However, considering the main contribution lies in the predictive advantage of CI, there is still room for further improvement in terms of overall model performance.
Formula~\ref{atttune} and~\ref{fixtune} illustrate the different process for the two tuning methods for the same input x.
\begin{align}
	X_{out}&=X_{in}\odot \operatorname{Attention}(WX_{in}) \label{atttune}\\
	X_{out}&=WX_{in}\label{fixtune}
\end{align}	
$\operatorname{Attention}$ implies obtaining coefficients (also called attention weights) that are specific to the temporal semantics of a single input $X_{in}$. The self-attention mechanism is widely used in LSTF to achieve this.
In the following section, we introduce our approach of using attention for special semantic modulation. What distinguishes our work is that we modulate both CI and CD, and address the limitations of using fixed-weights for simple modulation.
Also, by providing explicit evidence, we have demonstrated the effectiveness of attention in LSTF, addressing DLinear's question about the attention mechanism for LSTF.
To facilitate understanding without loss of generality, we consider an input combination of CI and CD denoted as $X = [x_{1}, x_{2}, x_{3}] \in \mathbb{R}^{1 \times 3}$. Here, $x_1$ and $x_2$ represent the output of CI and $x_3$ represents the output of CD. The process of tuning CI and CD using a fixed weight $W\in \mathbb{R}^{3\times 3}$ is described as follows:
\begin{equation}
	\begin{split}
		\begin{bmatrix}
			x_{1} & x_{2} & x_{3} \\
		\end{bmatrix}
		\begin{bmatrix}
			w_{11} & w_{12} & w_{13} \\
			w_{21} & w_{22} & w_{23} \\
			w_{31} & w_{32} & w_{33}
		\end{bmatrix}=
		\begin{bmatrix}
			\sum_{i=1}^{3} w_{i1} x_i, &\sum_{i=1}^{3} w_{i2} x_i, &\sum_{i=1}^{3}  w_{i3} x_i
		\end{bmatrix}
	\end{split}
	\label{matrixmulti}
\end{equation}
The drawback of this approach is that, since weight $W$ is fixed, the combination pattern of different $X$ remains the same. For instance, consider $x_1=0.1$, $x_2=0.3$, $x_3=0.6$, and aggregation weights $w_{11}=1$, $w_{22}=2$, $w_{33}=3$. The sum is equal to $(1 \times 0.1 + 2 \times 0.3 + 3 \times 0.6)$, indicating that larger weights correspond to larger CD values. However, if the temporal semantics of the input time-series sequence change, for example, by swapping the values of $x_1$ and $x_3$, this means that larger values are assigned smaller weights.

By making effective use of attention mechanisms, it is possible to tune the combination of CI and CD based on the temporal dependencies of different input sequences. Simplifying the result of Equation~\ref{matrixmulti} as $Q = [w_{:1}X, w_{:2}X, w_{:3}X]$, and defining another weight combination as $K= [t_{:1}X, t_{:2}X, t_{:3}X]$, the process of discovering temporal context using $QK$ can be described as follows:
\begin{equation}
	\begin{split}
		\begin{bmatrix}
			w_{:1} X &w_{:2} X &w_{:3} X 
		\end{bmatrix}
	\begin{bmatrix}
		t_{:1} X \\
		t_{:2} X \\
		t_{:3} X 
	\end{bmatrix}=
	\begin{bmatrix}
		 t_{:1}w_{:1} X^2+ t_{:2}w_{:2} X^2+ t_{:3}w_{:3} X^2
	\end{bmatrix}
	\end{split}
	\label{}
\end{equation}
Upon closer examination of the combination of $QK$, the presence of $X^2$ causes the coefficients to vary according to the temporal semantics of the input sequence. The introduction of the square operation, as well as the tuning of dual weights, enables the automatic tuning of attention to semantic information with different weights.
The coefficients obtained from input $X$ and possessing special semantic meanings are defined as $context$. 
The difference between using fixed-weights and TSA-weights for aggregation is as follows:
\begin{equation}
	\begin{split}
		context\cdot
		\begin{bmatrix}
			x_1 &x_2&x_3 
		\end{bmatrix}W\quad \leftrightarrow\quad
	\begin{bmatrix}
		x_1 &x_2&x_3 
	\end{bmatrix}W
	\end{split}
	\label{}
\end{equation}
\subsubsection{Attention modulation for CI and CD.}
%
%
We employ efficient-attention~\cite{shen2021efficient} to model the sequence due to its linear time complexity and effectiveness, as follows:
\begin{align}
	&E(Q,K,V) = (KV^{\top})^{\top}Q\\
	\text{where},K&=\rho_{k}(X),V=\rho_{v}(X)~\text{and}~ Q=\rho_{q}(X)\\
	\text{and},K &\in \mathbb{R}^{d_{k}\times d},V \in \mathbb{R}^{d_{v}\times d} \text{and}~Q \in \mathbb{R}^{d_{q}\times d}.
\end{align}
There are two options available for selecting $\rho_{i}(\cdot)$.
If convolutional mapping is used, then $Q$, $K$, and $V$ represent the modeled sequence of $X$. 
On the other hand, if linear mapping is employed, then $Q$, $K$, and $V$ represent the enriched features of a single sequence of $X$. 
The intuitive interpretation of these two approaches is illustrated as:
%
\begin{equation}
	\begin{split}
		W_{conv}\begin{bmatrix}
			x_{1} & x_{2} & x_{3} \\
			y_{1} & y_{2} & y_{3} \\
			z_{1} & z_{2} & z_{3}
		\end{bmatrix}
		\quad \leftrightarrow\quad &
		W_{linear}\begin{bmatrix}
			x_{1} & y_{1} & z_{1} \\
			x_{2} & y_{2} & z_{2} \\
			x_{3} & y_{3} & z_{3}
		\end{bmatrix}
	\end{split}
	\label{matrix}
\end{equation}
where $X=[x_1, x_{2}, x_{3} ]$, $Y=[y_{1}, y_{2}, y_{3}]$, and $Z=[z_{1}, z_{2}, z_{3}]$ represent different sequences, and convolutional weights $W_{conv}$ interact with different sequences. Linear weights $W_{linear}$ focus on enriching the features of a single sequence. However, in experiments, we did not observe significant differences between these two approaches.
\subsection{Deep Supervision for CI and CD}
Attention-based mechanisms can adjust a model's adaptation to inputs with different semantic features by incorporating the semantics of the input features. However, it has been observed that the current state-of-the-art approaches of this method often require stacking multiple layers to achieve performance improvement, which leads to significant computational overhead. Hence, we propose a novel optimization perspective that focuses on the model's deep parameters. By fine-tuning these parameters, we are able to improve the model's performance while reducing the computational cost. This approach permits the model to achieve better performance without the need for extensive computational resources.

Referring to the CI and CD components on the left side of Figure~\ref{fig:example}, their outputs $X_\text{CI}$ and $X_\text{CD}$ are both of $\in \mathbb{R}^{S\times n}$. 
From an abstract perspective, they can be seen as two separate sequence predictors, as shown in Figure~\ref{fig:example}. 
As they are far away from the output layer, we propose to improve the fitting of deep parameters for the two independent sequence predictors by separately supervising the CI and CD components at a deep level. Specifically, given the true prediction sequence y and the output of the two individual predictors $X_{\text{CI}}$ and $X_{\text{CD}}$, the process is defined as follows:
\begin{align}
	Distance_{\text{CI}} &= Loss(X_{\text{CI}},Y_{true})\\
	Distance_{\text{CD}} &= Loss(X_{\text{CD}},Y_{true})
\end{align}
where $Distance_{\text{CI}}$ and $Distance_{\text{CD}}$ represent the errors between the predicted results $X_{\text{CI}}$ and $X_{\text{CD}}$ of the CI and CD predictors and the true results $Y_{true}$. Our objective is to minimize $Distance_{\text{CI}}$ and $Distance_{\text{CD}}$. From a more intuitive perspective, if the independent prediction results are both highly accurate, then it is more likely that the joint prediction result will also be accurate.

Define the ultimate prediction target as: 
\begin{equation}
	X_{CD+CI} = \operatorname{Process}(X_{CD}, X_{CI})
\end{equation}
where $\operatorname{Process}(\cdot, \cdot)$ refers to the processing procedure after incorporating CI and CD.

Therefore, the optimization target $Distance_{\text{CI+CD}}$ generated by MLinear's final prediction is:
\begin{equation}
	Distance_{\text{CI+CD}} = Loss(X_{\text{CI+CD}},Y_{true})
\end{equation}
\subsection{Loss Function}
While previous models have used mean squared error (MSE) as the loss function, our experiments have demonstrated that Huber loss and mean absolute error (MAE) can achieve better performance. Huber loss exhibits greater robustness to outliers than MSE, allowing it to better handle outliers and improve model robustness. MAE imposes a relatively smaller penalty on outliers compared to MSE, leading to faster convergence during model training. 

Empirically, we propose a novel loss function that combines the advantages of MAE and Huber loss. It is worth noting that we did not carefully adjust the fusion weights, and only used a simple addition method. Nevertheless, in most tests, it demonstrated significant performance improvements. The trade-off between different loss functions will be left for future work.
The new loss function we propose is shown below:
\begin{equation}
	{L_\sigma }(y,f(x)) = \left\{ {\begin{array}{*{20}{l}}
			{\frac{1}{2}{{(y - f(x))}^2} + |y - f(x)|,}&{|y - f(x)| \le \sigma }\\
			{(\sigma  + 1)|y - f(x)| - \frac{1}{2}{\sigma ^2},}&{|y - f(x)| > \sigma }
	\end{array}} \right.
\end{equation}
Taking into account the separate prediction results for CI and CD, the total loss of the Mlinear can be expressed as:
\begin{equation}
	Loss = {L_\sigma }(Y_{true},f(X_{CD})) + {L_\sigma }(Y_{true},f(X_{CI})) + {L_\sigma }(Y_{true},f(X_{CI+CD}))
\end{equation}
It can be seen from the objective function that deep supervision is implemented to the hidden layers in the Mlinear to
generate more distinguish features. 
While for testing process, only $X$ will be used to obtain the final forecasting result.

\subsection{Model analysis}
We discussed the differences between Mlinear and two representative time-series prediction methods - the latest Transformer-based approach, PatchTST, and the non-Transformer method, Dlinear - with the aim of highlighting the unique features of our design concept.
\subsubsection{Comparation with PatchTST}
PatchTST is a Transformer-based time-series prediction method. Its main design concept is to explore the application of channel attention in Transformers and to use patches to explore local temporal dependencies, leveraging the concept of local semantics used in vision Transformers. We have summarized the differences between our design concept and PatchTST. Firstly, with regard to the use of attention mechanisms, PatchTST aims to model local temporal dependencies between channel-independent sequences using attention mechanisms. In contrast, our attention mechanism is designed to jointly model the temporal semantics of individual input time series for both channel information (CI) and contextual information (CD). Secondly, with regard to the utilization of CI and CD, PatchTST's main contribution is the introduction of CI into the Transformer model. In comparison, we fully utilize the advantages of CI and CD and cleverly combine them.
\subsubsection{Comparation with Dlinear}
Dlinear is a non-Transformer, simple linear combination time-series prediction method. The main idea of Dlinear is to use a simple linear layer to predict a single sequence of CI. In comparison, firstly, as with the differences between our method and PatchTST, our approach fully utilizes the joint prediction advantage of CI and CD, which Dlinear does not have. Secondly, from a model design perspective, Dlinear is conceptually simple, and its performance mainly comes from the full utilization of CI properties. In contrast, we cleverly combine the advantages of the Transformer and Linear models, using Linear to obtain joint predictions of CI and CD, and using Transformer to adaptively adjust the weighting of CI and CD under different temporal semantics.
\subsubsection{Time complexity analysis}
From Figure~\ref{fig:example}, it can be observed that mlinear mainly consists of two different types of layers, namely Linear and Attention layers. 
When considering the first $n$ linear layers, the all weight of them are $(L, S)$. Since the dimension of the input $(X_{1}, ..., X_{n})$ are all $(L, v)$, the multiplication complexity of a single linear layer and input $X_i$ is $O(vLS)$.

For the $(n+1)$-th linear layer, since the input $X_{n+1}$ is $(L, v)$ and the weight of $(n+1)$-th linear individual is $(L, S)$, the time complexity of the $(n+1)$-th linear layer is $O(LS)$.

As for the $(n+2)$-th linear layer, since the input is a concatenation of CI and CD, the input sequence X is $(2S, v)$, while the weight of $(n+2)$-th linear layer is $(2S, S)$. 
Therefore, the time complexity of the $(n+2)$-th layer is $O(2S^{2})$.

Condersing the Efficient-Attention for tune CI and CD, we introduce efficient-Attention to calculate the $KVQ$ order, reducing the time complexity from the quadratic time complexity of self-attention based on the sequence input length to $O(d_kd_v)$, which reduces the time to linear level. In the specific implementation of this paper, $d_k=\frac{S}{P}$, $d_q=d_k$, $d_v=2$, and the time complexity is significantly lower than that of traditional Transformer-based models.

\section{Experiments}
\label{sec:exp}

\subsection{Experimental Setup}\label{sec:setup}
\subsubsection{Datasets}
We conducted tests on seven datasets, including power, weather, and disease, which have been widely used as benchmarks in recent years.
Apart from the ETT dataset series collected by Informer~\cite{zhou2021informer}, all other datasets can be found in Autoformer~\cite{wu2021autoformer}.
\begin{enumerate}
	\item ETT (Electricity Transformer Temperature)~\cite{zhou2021informer} dataset consists of four datasets at two different granularities. The datasets include hourly data (ETTh1, ETTh2) and 15-minute data (ETTm1, ETTm2). Each dataset covers a two-year period from July 2016 to July 2018 and contains seven features related to oil and load for electricity transformers.
	\item Weather includes 21 indicators of weather, such as air temperature, and humidity. Its data is recorded every 10 min for 2020 in Germany.
	\item Electricity records the hourly electricity consumption of 321 clients between 2012 and 2014. 
	\item ILI measures the ratio of patients with influenza-like illness to the total number of patients, and includes weekly data from the Centers for Disease Control and Prevention in the United States from 2002 to 2021.
\end{enumerate}
\subsubsection{Setting}
To ensure fair comparison, we use the same batch size as PatchTST~\cite{nie2022time} on each dataset. 
All models follow the same experimental setup, with the prediction length T~$\in \{24, 36, 48, 60\}$ for the ILI dataset and T~$\in \{96, 192, 336, 720\}$ for other datasets as in the original PatchTST paper. 
We collect baseline results from the PatchTST paper, as they have set a strict standard to avoid underestimating the baseline. We use MSE and MAE as metrics for multivariate time series forecasting. For MLinear, following PatchTST, we use two versions of the lookback window, 336 and 512, to provide a strict comparison, and we label the model name with MLinear (336 or 512) during the quatitative comparison.
Apart from quantitative experiments, we used MLinear~(336) for all other experiments. For the sake of convenience in naming, we use ``MLinear'' without the suffix ``336'' in all other experiments.
The loss functions used are represented as follows: ``L1-Loss'' refers to MAE (Mean Absolute Error) Loss, ``L2-Loss'' refers to MSE (Mean Squared Error) Loss, ``H-Loss'' refers to Huber Loss, and "M-Loss" refers to the loss function proposed in our work. 
\subsubsection{Baselines}
The baseline methods that we compared cover the most outstanding and latest time-series prediction methods currently available. 
In particular, we emphasize PatchTST~\cite{nie2022time} and DLinear~\cite{zeng2022transformers}, as they are currently the most outstanding Transformer-based and non-Transformer-based LSTF methods, respectively.
\begin{enumerate}
	\item PatchTST~\cite{nie2022time} is a transformer-based time series prediction model that divides time series data into multiple small blocks (patches) and employs channel-independent (CI) input encoding for time series.
	\item DLinear~\cite{zeng2022transformers} takes full advantage of the channel-independent property, and achieves superior prediction performance beyond most Transformer-based models by simply processing single-channel time series through a single linear layer.
	\item Autoformer~\cite{wu2021autoformer} first applies seasonal-trend decomposition behind each neural block, which is a standard method in time series analysis to make raw data more predictable. 
	\item FEDformer~\cite{zhou2022fedformer} proposes the mixture of experts' strategy on top of the decomposition scheme of Autoformer, which combines trend components extracted by moving average kernels with various kernel sizes.
	\item Informer~\cite{zhou2021informer} analyzes attention mechanism in time series prediction, proposes a sparse attention mechanism to save computation and time costs, and achieves improvements in both efficiency and prediction accuracy compared to Transformer by incorporating feature distillation and one-step time series prediction.
	\item Pyraformer~\cite{liu2021pyraformer} is a Transformer-based model for time series forecasting. It can capture multi-resolution features and time dependencies, and achieves low time and space complexity while capturing long-term patterns.
	\item LogTrans's~\cite{li2019enhancing} core component is the causal convolutional self-attention, which generates queries and keys through causal convolutions to better incorporate local context into the attention mechanism.	
\end{enumerate}

%

%
%
%
\subsection{Performance Comparison and Analysis}
\begin{table*}[t]
	\centering
	\footnotesize
	\caption{Multivariate long-term forecasting results with supervised PatchTST. We use prediction lengths T $\in$ \{24, 36, 48, 60\} for ILI dataset and T $\in$ \{96, 192, 336, 720\} for the others. The best results are in bold and the second-best are underlined. In the count rows, the first number represents the number of times with the best performance, and the second number represents the number of times with the second-best performance.}
	\label{results}
	\resizebox{1\columnwidth}{!}{
		\begin{tabular}{cc|cc|cc|cc|cc|cc|cc|cc|cc|cc|cc}
			\toprule
			\multicolumn{2}{c|}{\multirow{2}{*}{{Models}}}&\multicolumn{2}{c|}{\multirow{2}{*}{{MLinear(336)}}} &    \multicolumn{2}{c|}{\multirow{2}{*}{{MLinear(512)}}}                              & \multicolumn{2}{c|}{PatchTST(336)} &\multicolumn{2}{c|}{PatchTST(512)}& \multicolumn{2}{c|}{DLinear} & \multicolumn{2}{c|}{FEDformer} & \multicolumn{2}{c|}{Autoformer} & \multicolumn{2}{c|}{Informer} & \multicolumn{2}{c|}{Pyraformer} & \multicolumn{2}{c}{LogTrans} \\
			
			\multicolumn{2}{c|}{}&\multicolumn{2}{c|}{} &    \multicolumn{2}{c|}{}                            & \multicolumn{2}{c|}{\textbf{ICLR 2023}} &\multicolumn{2}{c|}{\textbf{ICLR 2023}}& \multicolumn{2}{c|}{\textbf{AAAI 2023}} & \multicolumn{2}{c|}{\textbf{ICML 2022}} & \multicolumn{2}{c|}{\textbf{NIPS 2022}} & \multicolumn{2}{c|}{\textbf{AAAI 2021}} & \multicolumn{2}{c|}{\textbf{ICLR 2022}} & \multicolumn{2}{c}{\textbf{NIPS 2019}} \\
			\midrule
			\multicolumn{2}{c|}{Metric}& MSE& MAE& MSE& MAE&MSE& MAE&MSE& MAE& MSE& MAE& MSE& MAE& MSE& MAE& MSE& MAE& MSE& MAE& MSE& MAE\\ \midrule
			
			\multicolumn{1}{c|}{\multirow{4}{*}{\rotatebox{90}{Weather}}}  & 96&\uline{0.146}&	\uline{0.188}& \textbf{0.143}&	\textbf{0.186}

			& 0.152& 0.199&0.149 &0.198& 0.176 & 0.237 & 0.238& 0.314& 0.249& 0.329& 0.354& 0.405& 0.896          & 0.556 & 0.458& 0.490\\
			
			\multicolumn{1}{c|}{}&192&\uline{0.188}&\uline{0.231}	& \textbf{0.185}&\textbf{0.230}

			& 0.197           & 0.243    &0.194  &0.241      & 0.220         & 0.282        & 0.275          & 0.329         & 0.325          & 0.370          & 0.419         & 0.434         & 0.622          & 0.624          & 0.658         & 0.589        \\
			
			\multicolumn{1}{c|}{}& 336&\uline{0.240}&\uline{0.273}	& \textbf{0.236}&\textbf{0.271}

			& 0.249           & 0.283   &0.245& 0.282       & 0.265         & 0.319        & 0.339          & 0.377         & 0.351          & 0.391          & 0.583         & 0.543         & 0.739          & 0.753          & 0.797         & 0.652        \\
			\multicolumn{1}{c|}{}& 720&\uline{0.314}&\textbf{0.322}		&\textbf{0.307}&	\uline{0.325}

			& 0.320           & 0.335    &\uline{0.314} & 0.334     & 0.323         & 0.362        & 0.389          & 0.409         & 0.415          & 0.426          & 0.916         & 0.705         & 1.004          & 0.934          & 0.869         & 0.675        \\ \midrule
			\multicolumn{1}{c|}{\multirow{4}{*}{\rotatebox{90}{Electricity}}} & 96  &0.133&	0.225&	0.131&	0.223

			& \uline{0.130}          & \uline{0.222}    &\textbf{0.129} &\textbf{0.222}     & 0.140         & 0.237        & 0.186          & 0.302         & 0.196          & 0.313          & 0.304         & 0.393         & 0.386          & 0.449          & 0.258         & 0.357        \\
			
			\multicolumn{1}{c|}{}& 192  &0.149&\uline{0.239}	&	\textbf{0.146}&\textbf{0.237}

			& 0.148           & 0.240    &\uline{0.147} &0.240      & 0.153         & 0.249        & 0.197          & 0.311         & 0.211          & 0.324          & 0.327         & 0.417         & 0.386          & 0.443          & 0.266         & 0.368        \\
			\multicolumn{1}{c|}{}& 336&0.165&	\uline{0.255}&	\textbf{0.162}&	\textbf{0.253}

			& 0.167           & 0.261   &\uline{0.163} &0.259       & 0.169         & 0.267        & 0.213          & 0.328         & 0.214          & 0.327          & 0.333         & 0.422         & 0.378          & 0.443          & 0.280         & 0.380        \\
			\multicolumn{1}{c|}{}& 720 &0.203&\uline{0.288}&	\uline{0.199}&	\textbf{0.286}

			& {0.202}  & 0.291   &\textbf{0.197} &0.290       & 0.203         & 0.301        & 0.233          & 0.344         & 0.236          & 0.342          & 0.351         & 0.427         & 0.376          & 0.445          & 0.283         & 0.376        \\ \midrule
			
			\multicolumn{1}{c|}{\multirow{4}{*}{\rotatebox{90}{ILI}}}& 24&1.564	&\underline{0.762}&	1.564&	\underline{0.762}

			& \underline{1.522}           & 0.814    &\textbf{1.319} &  \textbf{0.754}    & 2.215         & 1.081        & 2.624          & 1.095         & 2.906          & 1.182          & 4.657         & 1.449         & 1.420          & 2.012          & 4.480         & 1.444        \\

			\multicolumn{1}{c|}{}& 36 &1.543&	\textbf{0.769}&	1.543	&\textbf{0.769}

			& \textbf{1.430}           & 0.834    &1.579 &0.870      & 1.963         & 0.963        & 2.516          & 1.021         & 2.585          & 1.038          & 4.650         & 1.463         & 7.394          & 2.031          & 4.799         & 1.467        \\
			
			\multicolumn{1}{c|}{}& 48&1.712&	\underline{0.823}&	1.712&	\underline{0.823}

			& \underline{1.673}           & 0.854  &\textbf{1.553} & \textbf{0.815}       & 2.130         & 1.024        & 2.505          & 1.041         & 3.024          & 1.145          & 5.004         & 1.542         & 7.551          & 2.057          & 4.800         & 1.468        \\

			\multicolumn{1}{c|}{} & 60&1.710&	\underline{0.834} &	1.710&	\underline{0.834}

			& \underline{1.529}          & 0.862  &\textbf{1.470} &\textbf{0.788}        & 2.368         & 1.096        & 2.742          & 1.122         & 2.761          & 1.114          & 5.071         & 1.543         & 7.662          & 2.100          & 5.278         & 1.560        \\ \midrule

			\multicolumn{1}{c|}{\multirow{4}{*}{\rotatebox{90}{ETTh1}}}& 96&\textbf{0.359}&	\textbf{0.385}&	\uline{0.362}&	\uline{0.388}

			& 0.375           & 0.399  &0.370& 0.400        & 0.375         & 0.399        & 0.376          & 0.415         & 0.435          & 0.446          & 0.941         & 0.769         & 0.664          & 0.612          & 0.878         & 0.740        \\
			
			\multicolumn{1}{c|}{}& 192   &\textbf{0.400}&	\textbf{0.412}&	\uline{0.402}&	\uline{0.414}

			& 0.414           & 0.421  &0.413&  0.429       & 0.405         & 0.416        & 0.423          & 0.446         & 0.456          & 0.457          & 1.007         & 0.786         & 0.790          & 0.681          & 1.037         & 0.824        \\
			
			\multicolumn{1}{c|}{}& 336   &\textbf{0.417}&\textbf{0.421}	&	\textbf{0.417}&	\uline{0.423}

			& 0.431           & 0.436  &\uline{0.422}& 0.440        & 0.439         & 0.443        & 0.444          & 0.462         & 0.486          & 0.487          & 1.038         & 0.784         & 0.891          & 0.738          & 1.238         & 0.932        \\
			
			\multicolumn{1}{c|}{}& 720&\textbf{0.429}&	\textbf{0.448}&	\uline{0.430}&\uline{0.450}

			& 0.449           & 0.466   &0.447&  0.468      & 0.472         & 0.490        & 0.469          & 0.492         & 0.515          & 0.517          & 1.144         & 0.857         & 0.963          & 0.782          & 1.135         & 0.852        \\ \midrule
			
			\multicolumn{1}{c|}{\multirow{4}{*}{\rotatebox{90}{ETTh2}}}& 96&0.274&\textbf{0.331}	&\textbf{0.271}	&\textbf{0.331}

			& \uline{0.274}           & \uline{0.336}    &\uline{0.274} &0.337      & 0.289         & 0.353        & 0.332          & 0.374         & 0.332          & 0.368          & 1.549         & 0.952         & 0.645          & 0.597          & 2.116         & 1.197        \\
			
			\multicolumn{1}{c|}{}& 192&\uline{0.334}&	\uline{0.372}&	\textbf{0.332}&	\textbf{0.370}

			& 0.339           & 0.379    &0.341 &0.382      & 0.383         & 0.418        & 0.407          & 0.446         & 0.426          & 0.434          & 3.792         & 1.542         & 0.788          & 0.683          & 4.315         & 1.635        \\

			\multicolumn{1}{c|}{} & 336&0.332&	\uline{0.381}&\textbf{0.329}	&\uline{0.381}

			& \uline{0.331}           & \textbf{0.380}&0.329& 0.384 & 0.448         & 0.465        & 0.400          & 0.447         & 0.477          & 0.479          & 4.215         & 1.642         & 0.907          & 0.747          & 1.124         & 1.604        \\
			
			\multicolumn{1}{c|}{}& 720&0.377&	\textbf{0.413}&	\textbf{0.375}&	\textbf{0.414}

			& 0.379           & 0.422   &0.379& 0.422       & 0.605         & 0.551        & 0.412          & 0.469         & 0.453          & 0.490          & 3.656         & 1.619         & 0.963          & 0.783          & 3.188         & 1.540        \\ \midrule
			
			\multicolumn{1}{c|}{\multirow{4}{*}{\rotatebox{90}{ETTm1}}}& 96&\textbf{0.289}&	\textbf{0.329}&	\uline{0.291}&	\uline{0.333}

			& \textbf{0.290}  & 0.342     &0.293 & 0.346    & 0.299         & 0.343        & 0.326          & 0.390         & 0.510          & 0.492          & 0.626         & 0.560         & 0.543          & 0.510          & 0.600         & 0.546        \\
			
			\multicolumn{1}{c|}{}& {192} &\textbf{0.331}&	\textbf{0.354}&	0.333&	\uline{0.356}

			& \uline{0.332}  & 0.369    &0.333 &0.370      & 0.335         & 0.365        & 0.365          & 0.415         & 0.514          & 0.495          & 0.725         & 0.619         & 0.557          & 0.537          & 0.837         & 0.700        \\

			\multicolumn{1}{c|}{}& {336} &0.369&	\textbf{0.376}&	\textbf{0.364}	&\textbf{0.377}
			
			& \uline{0.366}  & 0.392      &0.369& 0.392    & 0.369         & \uline{0.386}        & 0.392          & 0.425         & 0.510          & 0.492          & 1.005         & 0.741         & 0.754          & 0.655          & 1.124         & 0.832        \\

			\multicolumn{1}{c|}{}& {720}&0.428&	\textbf{0.410}&	0.422	&{0.409}
			
			& \uline{0.420}  & 0.424  &\textbf{0.416} & \uline{0.420}       & 0.425         & 0.421        & 0.446          & 0.458         & 0.527          & 0.493          & 1.133         & 0.845         & 0.908          & 0.724          & 1.153         & 0.820        \\ \midrule
			
			\multicolumn{1}{c|}{\multirow{4}{*}{\rotatebox{90}{ETTm2}}}& {96}&\uline{0.159}&\textbf{0.243}	&	\textbf{0.159}&	\textbf{0.243}
			& 
			
			0.165           & \uline{0.255} &0.166 &0.256         & 0.167         & 0.260        & 0.180          & 0.271         & 0.205          & 0.293          & 0.355         & 0.462         & 0.435          & 0.507          & 0.768         & 0.642        \\
			
			\multicolumn{1}{c|}{}& 192&\uline{0.215}&	\uline{0.283}&	\textbf{0.214}&\textbf{0.282}

			& 0.220           & 0.292   &0.223& 0.296       & 0.224         & 0.303        & 0.252          & 0.318         & 0.278          & 0.336          & 0.595         & 0.586         & 0.730          & 0.673          & 0.989         & 0.757        \\
			\multicolumn{1}{c|}{} & 336&\uline{0.272}&	\uline{0.320}&	\textbf{0.266}	&\textbf{0.316}

			& 0.278           & 0.329   &0.274 & 0.329      & 0.281         & 0.342        & 0.324          & 0.364         & 0.343          & 0.379          & 1.270         & 0.871         & 1.201          & 0.845          & 1.334         & 0.872        \\
			
			\multicolumn{1}{c|}{} & 720 &\textbf{0.351}&	\textbf{0.371}&	\uline{0.357}&	\uline{0.376}

			& {0.367}  & 0.385   &0.362 &0.385       & 0.397         & 0.421        & 0.410          & 0.420         & 0.414          & 0.419          & 3.001         & 1.267         & 3.625          & 1.451          & 3.048         & 1.328        \\ \midrule
			\multicolumn{2}{c|}{Count} &21&21&29& 18                                        & 3              & 12             & 10            & 6              & 0             & 1              & 0              & 0             & 0             & 0              & 0              & 0             & 0  &0& 0    &0     \\ \bottomrule
	\end{tabular}}
\end{table*}
\subsubsection{Quantitative experiment}
%
The comparative results on seven widely used long sequence prediction datasets are presented in Table~\ref{results}. Consistent with the latest baseline method PatchTST, we use two versions of MLinear for fair comparison, named as MLinear (input sequence length) with input sequence lengths of 336 and 512, respectively.
From Table~\ref{results}, we can see that both versions of MLinear, with input sequence lengths of 336 and 512, respectively, achieve significant prediction advantages over other baseline methods. 

Specifically, MLinear (512) achieves the most optimal performance 29 times and the second-best performance 18 times, which is significantly better than PatchTST (512) with the same input sequence length, which only achieves 10 and 6 times, respectively, and even outperforms MLinear (336). When the input sequence is limited to 336, MLinear maintains a significant performance lead over PatchTST, achieving the most optimal and second-best performance 21 times each, which is significantly better than PatchTST (336) with only 3 and 12 times, respectively.
Moreover, it is worth noting that due to the consideration of the characteristics of CI and CD, DLinear, PatchTST, and MLinear achieve a boundary-style lead over other comparative models. For example, on the weather dataset with the longest prediction length of 720, the performance gap between DLinear, PatchTST, and MLinear in terms of MSE is no more than 0.02. However, compared with the other best-performing Fedformer, the performance gap is nearly 0.08 (from MSE 0.389 by Fedformer to MSE 0.307 by Mlinear (512)).

\begin{table*}[htbp]
	\centering
	\resizebox{1\columnwidth}{!}{
		\begin{tabular}{cc|cc|cc|cc|cc||cc|cc|cc|cc}
			\toprule
			\multicolumn{2}{c|}{Loss}&\multicolumn{8}{c||}{L2-Loss}&\multicolumn{8}{c}{M-Loss}\\\midrule
			\multicolumn{2}{c|}{Metric} & MSE & MAE & MSE & MAE & MSE & MAE &MSE&MAE& MSE & MAE & MSE & MAE & MSE & MAE &MSE&MAE\\ \midrule
			\multicolumn{2}{c|}{Methods}&\multicolumn{2}{c|}{CD} & \multicolumn{2}{c|}{CI} & \multicolumn{2}{c|}{S-ATT} &\multicolumn{2}{c||}{D-ATT}&\multicolumn{2}{c|}{CD} & \multicolumn{2}{c|}{CI} & \multicolumn{2}{c|}{S-ATT} &\multicolumn{2}{c}{D-ATT}\\ \midrule
			
			\multicolumn{1}{c|}{\multirow{4}{*}{\rotatebox{90}{Electricity}}}&96 &0.140&	0.236&	0.134&	0.229	&0.134&	0.229&	\textbf{0.133}	&\textbf{0.228 }        &0.141&	0.234&	0.135&	\textbf{0.225}&	0.134&	\textbf{0.225}&	\textbf{0.133}&	\textbf{0.225}
			\\ 
			\multicolumn{1}{c|}{}&192 &0.154	&0.247&	\textbf{0.150}&	0.243	&\textbf{0.150}&	\textbf{0.242}&	\textbf{0.150}&	0.245			&0.154&	0.246	&0.150&	0.239&	\textbf{0.149}&	\textbf{0.239}&	\textbf{0.149}&	\textbf{0.239}
			\\ 
			\multicolumn{1}{c|}{}&336 &0.170&	0.264	&0.167&\textbf{0.260}&	0.167&	0.261&	\textbf{0.166}&	0.262 &0.170&	0.262&	0.167&	\textbf{0.255}&	0.166&	\textbf{0.255}&	\textbf{0.165}&	\textbf{0.255}
			
			\\ 
			\multicolumn{1}{c|}{}&720 & 0.209	&0.296&	0.206&\textbf{0.293}&	0.208	&0.296&	\textbf{0.204}&	0.295   &0.209&	0.294&	0.206&	0.288&	0.205&	\textbf{0.287}	&\textbf{0.203}&	0.288
			
			\\ \midrule
			
			\multicolumn{1}{c|}{\multirow{4}{*}{\rotatebox{90}{ETTh1}}}&96 & 0.370	&0.395&	0.378&	0.400	&0.371&	0.395&	\textbf{0.366}&	\textbf{0.394}    & 0.364	&0.387	&0.372	&0.392	&0.367&	0.389&	\textbf{0.359}&	\textbf{0.385}
			
			\\ 
			\multicolumn{1}{c|}{}&192 &0.407	&\textbf{0.415}&	0.416&	0.423&	0.415&	0.425&	\textbf{0.403}&	0.418  &0.406&	\textbf{0.410}	&0.413&	0.417&	0.407&	0.414&	\textbf{0.400}&	0.412
			\\ 
			\multicolumn{1}{c|}{}&336 & 0.424&	\textbf{0.425}&	0.429&	0.432	&0.431&	0.436&	\textbf{0.418}&	0.429  &0.429	&0.422	&0.433&	0.428&	0.428&	0.425&	\textbf{0.417}&	\textbf{0.421}
			\\ 
			\multicolumn{1}{c|}{}&720 & \textbf{0.432}	&\textbf{0.449}	&0.447&	0.456&	0.454&	0.464&	0.451&	0.465   & 0.425&	0.441&	0.440&	0.45&	0.434&	\textbf{0.447}&	\textbf{0.429}&	0.448
			
			\\ \midrule
			\multicolumn{1}{c|}{\multirow{4}{*}{\rotatebox{90}{ETTm1}}}&96 & 0.301&	0.343&	0.289&	0.336&	\textbf{0.288}&	\textbf{0.335}&	\textbf{0.288}&	0.337  & 0.297&	0.337	&0.292	&0.330&	0.291&	\textbf{0.329}&	\textbf{0.289}&	0.329
			
			\\ 
			\multicolumn{1}{c|}{}&192 & 0.338	&0.365	&0.331&	\textbf{0.360}&	0.331	&\textbf{0.360}&	\textbf{0.327}&	0.361    & 0.335&	0.359	&0.335&	0.355&	0.334	&\textbf{0.354}&	\textbf{0.331}&	\textbf{0.354}
			\\ 
			\multicolumn{1}{c|}{}&336 & 0.371	&0.384&	{0.370}&	\textbf{0.382}&	0.372&	0.385&	\textbf{0.364}&	0.383   & \textbf{0.369}&	0.379&	0.373	&0.377&	0.372&	\textbf{0.376}&	\textbf{0.369}	&0.376
			\\ 
			\multicolumn{1}{c|}{}&720 &\textbf{0.427}	&\textbf{0.416}	&0.429&	\textbf{0.416}&	0.432&	0.419&	{0.428}	&0.417  &\textbf{0.427}	&0.412&	0.431&	\textbf{0.411}&	0.430&	\textbf{0.411}&	0.428&	\textbf{0.410}
			\\ \midrule
			\multicolumn{2}{c|}{Count}&\multicolumn{2}{c|}{6}&\multicolumn{2}{c|}{6}&\multicolumn{2}{c|}{5}&\multicolumn{2}{c|}{12}&\multicolumn{2}{c|}{3}&\multicolumn{2}{c|}{3}&\multicolumn{2}{c|}{10}&\multicolumn{2}{c}{18}\\
			\bottomrule
			
	\end{tabular}}
	\caption{An intuitive comparison between CI, CD, CI and CD aggregation without deep supervision, and MLinear which with CI and CD aggregation with deep supervision. ``S-ATT'' refers to the use of a single loss MLinear without deep supervision, while ``D-ATT'' refers to the use of deep supervision with MLinear. The loss function used is M-Loss.}
	\vspace{-4cm}
	\label{CICDMIXDEEP}
\end{table*}
\begin{table}[htbp]
	\centering
	\scriptsize
	\begin{tabular}{cc|cc|cc|cc|cc}
		\toprule
		&&\multicolumn{2}{c|}{H-Loss} & \multicolumn{2}{c|}{M-Loss} & \multicolumn{2}{c|}{L1-Loss} & \multicolumn{2}{c}{L2-Loss} \\ 
		\midrule
		\multicolumn{2}{c|}{{{Metric}}} & MSE & MAE & MSE & MAE & MSE & MAE & MSE & MAE \\ \midrule
		\multicolumn{1}{c|}{\multirow{4}{*}{\rotatebox{90}{ETTh1}}}& 96& \underline{0.363} & 0.390 		& \textbf{0.359} & \underline{0.385} & \textbf{0.359} & \textbf{0.384} & 0.366 & 0.394 \\ 
		\multicolumn{1}{c|}{}&192& 0.402 & 0.415 		& \underline{0.400} & \underline{0.412} & \textbf{0.399} & \textbf{0.410} & 0.403 & 0.418 \\ 
		\multicolumn{1}{c|}{}&336& \textbf{0.417} & \underline{0.424} & \textbf{0.417} & \textbf{0.421} & 0.419 & \textbf{0.421} & \underline{0.418} & 0.429 \\ 
		\multicolumn{1}{c|}{}&720& 0.435 & 0.453 & \textbf{0.429} & \underline{0.448} & \underline{0.430} & \textbf{0.446} & 0.451 & 0.465 \\ \midrule
		\multicolumn{1}{c|}{\multirow{4}{*}{\rotatebox{90}{ETTh2}}}& 96& \textbf{0.274} & \underline{0.333} & \textbf{0.274} & \textbf{0.331} & \underline{0.276} & \textbf{0.331} & 0.283 & 0.340 \\ 
		\multicolumn{1}{c|}{}&192& 0.336 & \underline{0.375} & \textbf{0.334} & \textbf{0.372} & \underline{0.335} & \textbf{0.372} & 0.343 & 0.385 \\ 
		\multicolumn{1}{c|}{}& 336& 0.333 & 0.385 & \underline{0.332} & \underline{0.381} & \textbf{0.329} & \textbf{0.378} & 0.340 & 0.395 \\ 
		\multicolumn{1}{c|}{}& 720& 0.382 & \underline{0.419} & \textbf{0.377} & \textbf{0.413} & \underline{0.379} & \textbf{0.413} & 0.400 & 0.438 \\ \midrule
		\multicolumn{1}{c|}{\multirow{4}{*}{\rotatebox{90}{ETTm1}}}&  96& \underline{0.289} & \underline{0.334} & \underline{0.289} & \textbf{0.329} & 0.292 & \textbf{0.329} & \textbf{0.288} & 0.337 \\ 
		\multicolumn{1}{c|}{}& 192& \underline{0.329} & \underline{0.357} & 0.331 & \textbf{0.354} & 0.334 & \textbf{0.354} & \textbf{0.327} & 0.361 \\ 
		\multicolumn{1}{c|}{}&336& \underline{0.366} & \underline{0.379} & 0.369 & \textbf{0.376} & 0.372 & \textbf{0.376} & \textbf{0.364} & 0.383 \\ 
		\multicolumn{1}{c|}{}&720& \textbf{0.422} & \underline{0.413} & \underline{0.428} & \textbf{0.410} & 0.430 & \textbf{0.410} & \underline{0.428} & 0.416 \\ \midrule
		
		\multicolumn{1}{c|}{\multirow{4}{*}{\rotatebox{90}{ETTm2}}}& 96& \textbf{0.159}&	\underline{0.246}&	\textbf{0.159}&	\textbf{0.243}&	\underline{0.160}&	\textbf{0.243}&	0.162&	0.249 \\ 
		\multicolumn{1}{c|}{}&192 & \textbf{0.210}&	\textbf{0.280}&	\underline{0.215}&	0.283&	0.219&	\underline{0.282}&	0.219&	0.291 \\ 
		\multicolumn{1}{c|}{}&336& \textbf{0.261}&	\textbf{0.315}&	\underline{0.272}&	0.320&	0.274&	\underline{0.319}&	0.273&	0.325 \\ 
		\multicolumn{1}{c|}{}& 720& 0.357&	0.376&	\textbf{0.351}&	\underline{0.371}&	\underline{0.352}&	\textbf{0.370}&	0.369&	0.386 \\ \midrule
		\multicolumn{2}{c|}{Count}&  \multicolumn{1}{c}{8}&  \multicolumn{1}{c|}{13}&\multicolumn{1}{c}{17}&\multicolumn{1}{c|}{11}&\multicolumn{1}{c}{17}&\multicolumn{1}{c|}{8}&\multicolumn{1}{c}{3}&\multicolumn{1}{c}{2}\\ 
		\bottomrule
		
	\end{tabular}
	\caption{Performance results of mlinear using four different loss functions on four datasets.}
	\label{DIFFERENTLOSS}
\end{table}
\begin{table*}[htbp]
	\centering
	\begin{tabular}{cc|cccc|cccc}
		\toprule
		\multicolumn{2}{c|}{Metric}	 & MSE & MAE & MSE & MAE & MSE & MAE & MSE & MAE \\\midrule
		\multicolumn{2}{c|}{Loss}&\multicolumn{2}{c}{S-L1-Loss} & \multicolumn{2}{c|}{D-L1-Loss} & \multicolumn{2}{c}{S-L2-Loss} & \multicolumn{2}{c}{D-L2-Lloss}\\ \midrule
		\multicolumn{1}{c|}{\multirow{4}{*}{\rotatebox{90}{ETTh1}}}&92 & 0.379&0.402 &0.359($\downarrow$0.020)&0.384($\downarrow$0.018)&0.374&0.399&0.366($\downarrow$0.008)&0.394($\downarrow$0.005)
		\\ 
		\multicolumn{1}{c|}{}&192 & 0.414&0.420 &0.399($\downarrow$0.015)&0.410($\downarrow$0.010)&0.430&0.436&0.403($\downarrow$0.027)&0.418($\downarrow$0.018)
		\\ 
		\multicolumn{1}{c|}{}&336& 0.435&0.432 &0.419($\downarrow$0.016)&0.421($\downarrow$0.011)&0.452&0.456&0.418($\downarrow$0.034)&0.429($\downarrow$0.027)
		\\ 
		\multicolumn{1}{c|}{}&720& 0.450&0.463 &0.430($\downarrow$0.020)&	0.447($\downarrow$0.016)&	0.475&	0.482&	0.451($\downarrow$0.024)&	0.465($\downarrow$0.017)\\ \midrule
		
		\multicolumn{1}{c|}{\multirow{4}{*}{\rotatebox{90}{ETTh2}}}&92 & 0.282&0.333	&0.276($\downarrow$0.006)&	0.331($\downarrow$0.002)&	0.294&	0.350&	0.283($\downarrow$0.011)&	0.340($\downarrow$0.010)	\\ 
		\multicolumn{1}{c|}{}&192 & 0.359&0.381	&0.335($\downarrow$0.024)&	0.372($\downarrow$0.009)&	0.378&	0.402&	0.343($\downarrow$0.035)&	0.385($\downarrow$0.017)
		\\ 
		\multicolumn{1}{c|}{}&336 & 0.351&0.387	&0.329($\downarrow$0.022)&	0.378($\downarrow$0.009)&	0.353&	0.406&	0.340($\downarrow$0.013)&	0.395($\downarrow$0.009)
		\\ 
		\multicolumn{1}{c|}{}&720 & 0.375&0.412	&0.379($\downarrow$0.004)&	0.413(-0.001)&	0.411&	0.447&	0.400($\downarrow$0.011)&	0.438($\downarrow$0.009)
		\\ \midrule
		\multicolumn{1}{c|}{\multirow{4}{*}{\rotatebox{90}{ETTm1}}}&96 & 0.301&0.346	&0.292($\downarrow$0.009)&	0.329($\downarrow$0.017)&	0.305&	0.355&	0.288($\downarrow$0.017)&	0.337($\downarrow$0.018)
		\\ 
		\multicolumn{1}{c|}{}&192 & 0.342&0.367	&0.334($\downarrow$0.008)&	0.354($\downarrow$0.013)&	0.339&	0.372&	0.327($\downarrow$0.012)&	0.361($\downarrow$0.011)
		\\ 
		\multicolumn{1}{c|}{}&336 & 0.382&0.389	&0.372($\downarrow$0.010)&	0.376($\downarrow$0.013)&	0.382&	0.397&	0.364($\downarrow$0.018)&	0.383($\downarrow$0.014)
		\\ 
		\multicolumn{1}{c|}{}&720 & 0.444&0.424	&0.430($\downarrow$0.014)&	0.410($\downarrow$0.014)&	0.434&	0.430&	0.428($\downarrow$0.006)&	0.417($\downarrow$0.013)
		\\ \midrule
		
		\multicolumn{1}{c|}{\multirow{4}{*}{\rotatebox{90}{ETTm2}}}&96 & 0.163&0.243	&0.161($\downarrow$0.002)&	0.243($\downarrow$0.000)&	0.167&	0.255&	0.162($\downarrow$0.005)&	0.250($\downarrow$0.005)
		\\ 
		\multicolumn{1}{c|}{}&192 & 0.218&0.283	&0.219($\downarrow$0.001)&	0.283($\downarrow$0.000)&	0.234&	0.305&	0.219($\downarrow$0.015)&	0.291($\downarrow$0.014)
		\\ 
		\multicolumn{1}{c|}{}&336 & 0.287&0.326	&0.274($\downarrow$0.013)&	0.319($\downarrow$0.007)&	0.290&	0.344&	0.273($\downarrow$0.017)&	0.326($\downarrow$0.018)
		\\ 
		\multicolumn{1}{c|}{}&720 & 0.381&0.389	&0.352($\downarrow$0.011)&	0.370($\downarrow$0.019)&	0.382&	0.399&	0.370($\downarrow$0.012)&	0.386($\downarrow$0.013)
		\\ \midrule
		\multicolumn{2}{c|}{Loss}& \multicolumn{2}{c}{S-M-Loss} & \multicolumn{2}{c|}{D-M-Loss}& \multicolumn{2}{c}{S-H-Loss} & \multicolumn{2}{c}{D-H-Loss} \\\midrule
		\multicolumn{1}{c|}{\multirow{4}{*}{\rotatebox{90}{ETTh1}}}&92&0.380&0.403&0.359($\downarrow$0.021)&0.385($\downarrow$0.018)&0.385&0.408&	0.363($\downarrow$0.022)&	0.390($\downarrow$0.018)\\
		\multicolumn{1}{c|}{}&192&0.410&0.417&0.400($\downarrow$0.010)&0.412($\downarrow$0.005)&0.424&0.429&	0.402($\downarrow$0.020)&	0.415($\downarrow$0.014)\\
		\multicolumn{1}{c|}{}&336&0.442&0.440&0.417($\downarrow$0.025)&0.421($\downarrow$0.019)&0.443&0.443&0.417($\downarrow$0.026)&0.424($\downarrow$0.019)\\
		\multicolumn{1}{c|}{}&720&	0.459&0.468&0.429($\downarrow$0.030)&0.448($\downarrow$0.020)&0.460&0.470&0.435($\downarrow$0.025)&0.453($\downarrow$0.017)\\
		\midrule
		\multicolumn{1}{c|}{\multirow{4}{*}{\rotatebox{90}{ETTh2}}}&92&0.280&0.333&	0.274($\downarrow$0.006)&	0.331($\downarrow$0.002)&0.287	&0.342&	0.274($\downarrow$0.013)&	0.333($\downarrow$0.009)\\
		\multicolumn{1}{c|}{}&192&0.364&0.385&	0.334($\downarrow$0.030)&	0.372($\downarrow$0.013)& 0.366&	0.396&	0.336($\downarrow$0.030)&	0.375($\downarrow$0.021)\\
		\multicolumn{1}{c|}{}&336&	0.346&0.389	&0.332($\downarrow$0.014)&0.381($\downarrow$0.008)&0.348&	0.393&	0.332($\downarrow$0.016)&	0.385($\downarrow$0.008)\\
		\multicolumn{1}{c|}{}&720&	0.381&0.420&	0.377($\downarrow$0.004)&	0.413($\downarrow$0.007)& 0.383&	0.426&	0.382($\downarrow$0.001)&	0.419($\downarrow$0.007)\\
		\midrule
		\multicolumn{1}{c|}{\multirow{4}{*}{\rotatebox{90}{ETTm1}}}&92&	0.296&	0.343&	0.290($\downarrow$0.006)&0.330($\downarrow$0.013) &0.303&	0.352&	0.289($\downarrow$0.014)&	0.334($\downarrow$0.018)\\
		\multicolumn{1}{c|}{}&192&	0.343&	0.371&	0.331($\downarrow$0.012)&0.354($\downarrow$0.017)& 0.344&	0.374&	0.329($\downarrow$0.015)&	0.357($\downarrow$0.017)\\
		\multicolumn{1}{c|}{}&336&	0.386&	0.392&	0.369($\downarrow$0.017)&0.376($\downarrow$0.016)&0.384&	0.395&	0.366($\downarrow$0.018)&	0.379($\downarrow$0.016)\\
		\multicolumn{1}{c|}{}&720&	0.440&	0.421&	0.428($\downarrow$0.012)&0.410($\downarrow$0.011)&0.435&	0.422&	0.422($\downarrow$0.013)&	0.413($\downarrow$0.009)\\
		\midrule
		\multicolumn{1}{c|}{\multirow{4}{*}{\rotatebox{90}{ETTm2}}}&92&	0.158&	0.243&	0.159($\uparrow$0.001)&0.243(--0.000)& 0.165&	0.250&	0.160($\downarrow$0.005)&	0.246($\downarrow$0.004)\\
		\multicolumn{1}{c|}{}&192&	0.222&	0.287&	0.215($\downarrow$0.007)&0.283($\downarrow$0.004)& 0.225&	0.293&	0.211($\downarrow$0.014)&	0.281($\downarrow$0.012)\\
		\multicolumn{1}{c|}{}&336&	0.287&	0.330&	0.272($\downarrow$0.015)&0.320($\downarrow$0.010) &0.290&0.338&0.262($\downarrow$0.028)&	0.316($\downarrow$0.022)\\
		\multicolumn{1}{c|}{}&720&	0.379&	0.390&	0.351($\downarrow$0.028)	&0.371($\downarrow$0.019)&0.380&0.395&0.357($\downarrow$0.023)&0.377($\downarrow$0.018)\\
		\bottomrule
	\end{tabular}
	\caption{Ablation experiments of deep supervision, four different loss functions were used, with the prefix ``S-'' indicating simple mix experiments, which only used one final target loss, and the prefix ``D-'' indicating deep supervision experiments, which supervised CI, CD, and the final target.}
	\label{DEEPSUP}
\end{table*}

\begin{table*}[htbp]
	\centering
	\scriptsize
	\begin{tabular}{cc|cc||cc||cc}
		\toprule
		\multicolumn{2}{c|}{Methods} &\multicolumn{2}{c||}{3ATT} & \multicolumn{2}{c||}{Simple} & \multicolumn{2}{c}{1ATT}  \\ \midrule
		\multicolumn{2}{c|}{Metric} & MSE & MAE & MSE & MAE & MSE & MAE \\ \midrule
		\multicolumn{1}{c|}{\multirow{4}{*}{\rotatebox{90}{ETTh1}}}&92 & \textbf{0.359}&	\textbf{0.386}&	0.360&	\textbf{0.386}&	0.359&	\textbf{0.385}\\ 
		\multicolumn{1}{c|}{}&192& \textbf{0.399}&	\textbf{0.411}&	0.402&	0.412&	0.400	&0.412\\ 
		\multicolumn{1}{c|}{}&336& 0.418	&0.422	&0.422&	0.423&\textbf{0.417}&	\textbf{0.421}\\ 
		\multicolumn{1}{c|}{}&720& 0.435	&0.455	&\textbf{0.429}&	\textbf{0.445}&	\textbf{0.429}&	0.448\\ \midrule
		
		\multicolumn{1}{c|}{\multirow{4}{*}{\rotatebox{90}{ETTh2}}} &92& 0.277&	0.332&	0.277&	\textbf{0.331}&	\textbf{0.274}	&\textbf{0.331}\\ 
		\multicolumn{1}{c|}{} &192& \textbf{0.333}	&\textbf{0.372}&	0.335&\textbf{0.372}&	0.334	&\textbf{0.372}\\ 
		\multicolumn{1}{c|}{}&336 & 0.334	&0.382&	\textbf{0.331}&	\textbf{0.381}&	0.332&	\textbf{0.381}\\ 
		\multicolumn{1}{c|}{}&720 &\textbf{0.373}	&\textbf{0.411}&	0.377	&0.414&	0.377&	0.413\\ \midrule
		\multicolumn{1}{c|}{\multirow{4}{*}{\rotatebox{90}{ETTm1}}}&92 &0.291	&0.332&	0.291&	\textbf{0.330}&	\textbf{0.289}&	\textbf{0.329}	\\ 
		\multicolumn{1}{c|}{}&192& \textbf{0.331}	&0.355&	0.332&	\textbf{0.354}&	\textbf{0.331}&	\textbf{0.354}	\\ 
		\multicolumn{1}{c|}{}&336& 0.370	&0.377&	\textbf{0.369}&	\textbf{0.376}&	\textbf{0.369}&	\textbf{0.376}\\ 
		\multicolumn{1}{c|}{}&720&\textbf{0.425}&	\textbf{0.409}&	0.428&	0.410	&0.428&	0.410\\ \midrule
		
		\multicolumn{1}{c|}{\multirow{4}{*}{\rotatebox{90}{ETTm2}}}&92 & 0.161	&0.244&	0.161&	0.245&	\textbf{0.159}&	\textbf{0.243}\\ 
		\multicolumn{1}{c|}{}&192 &0.219&	\textbf{0.282}&	0.219&	0.285&	\textbf{0.215}&	0.283\\ 
		\multicolumn{1}{c|}{}&336& \textbf{0.271}	&\textbf{0.320}&	0.274&	0.323&	0.272&	\textbf{0.320}\\ 
		\multicolumn{1}{c|}{}&720 & 0.359	&0.374&	0.367&	0.381&	\textbf{0.351}&	\textbf{0.371}\\ \midrule
		\multicolumn{2}{c|}{Count}&\multicolumn{2}{c|}{14}&\multicolumn{2}{c|}{11}&\multicolumn{2}{c|}{20}\\\bottomrule
		
	\end{tabular}
	\caption{The comparison of three different aggregation methods for CI and CD includes ``Simple'', which means using a linear layer for aggregation, and ``1ATT'' and ``3ATT'', which mean using one or three layers of attention for aggregation, respectively. The loss function used is M-Loss.}
	\label{3AGG}
\end{table*}
\subsubsection{CI and CD mixed ablation experiment}
We compare the performance of four approaches using L2-Loss and M-Loss for training on datasets Electricity, ETTh1, and ETTm1 to demonstrate the superiority of deep supervision over CI and CD, as well as the simple combination of CI and CD.
From Table~\ref{CICDMIXDEEP}, it can be seen that the full version of MLinear achieves the best performance 12 times (with L2-Loss) and 18 times (with M-Loss) on the three datasets, while CI and CD achieve the best performance only 6 times (with L2-Loss) and 3 times (with M-Loss). The simple combination of CI and CD achieves the best performance 5 times (with L2-Loss) and 10 times (with M-Loss). This indicates that a simple combination of CI and CD alone cannot achieve a synergy effect greater than the sum of its parts.
%
\subsubsection{Loss function ablation experiment}
We compare the performance of MLinear using L2-Loss, L1-Loss, H-Loss, and M-Loss to highlight the superior performance of M-Loss. 

From Table~\ref{DIFFERENTLOSS}, it can be seen that the use of M-Loss achieves the best performance on ETTh1 three times with L2-Loss and once with L1-Loss. On ETTh2, it achieves the best performance three times with L2-Loss and three times with L1-Loss. On ETTm1, it achieves the best performance four times with L1-Loss. On ETTm2, it achieves the best performance twice with L2-Loss and once with L1-Loss.
\begin{figure*}[h]
	\centering
	\subfloat[]{\includegraphics[width=0.33\textwidth]{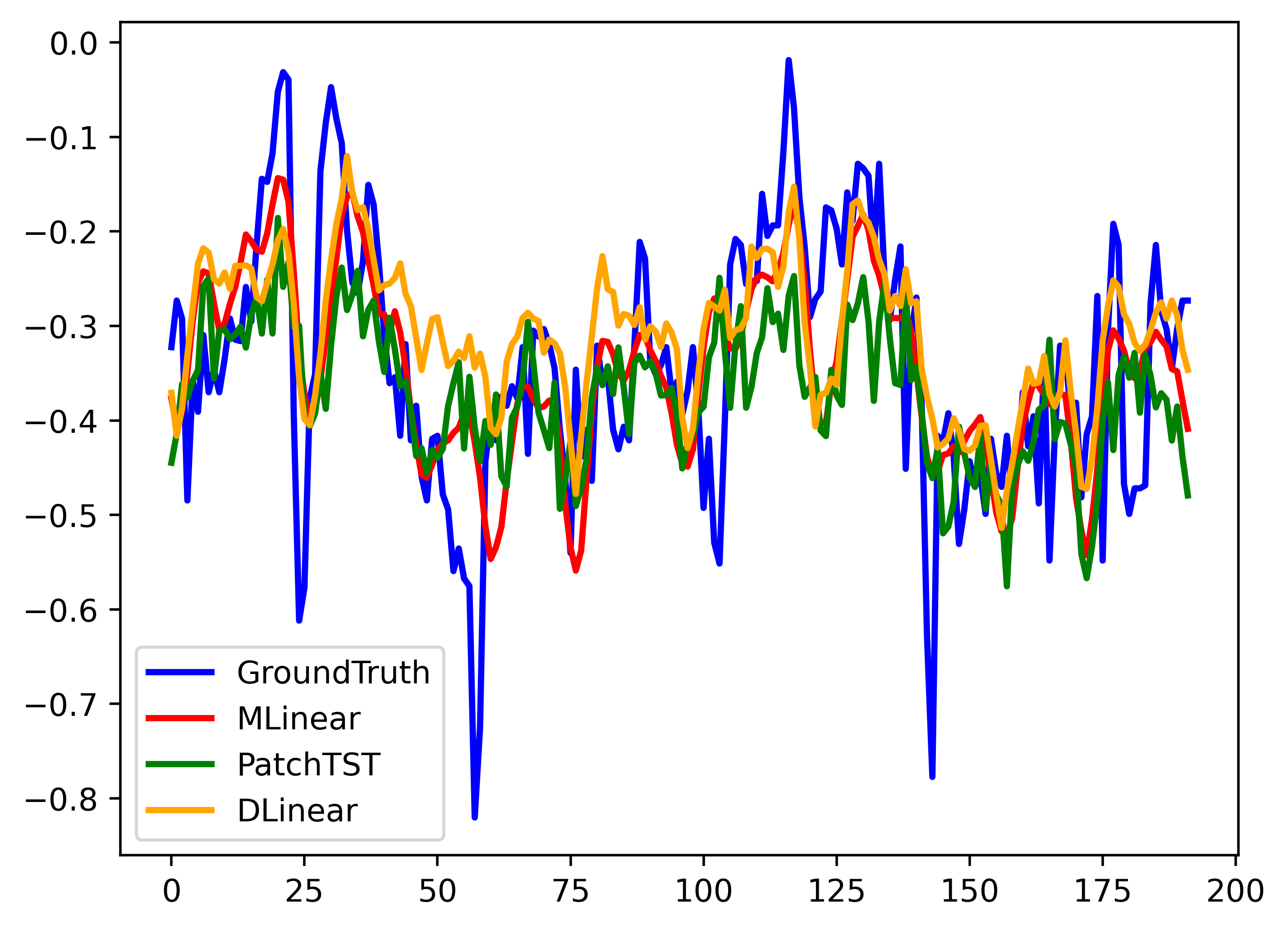}}
	\subfloat[]{\includegraphics[width=0.3175\textwidth]{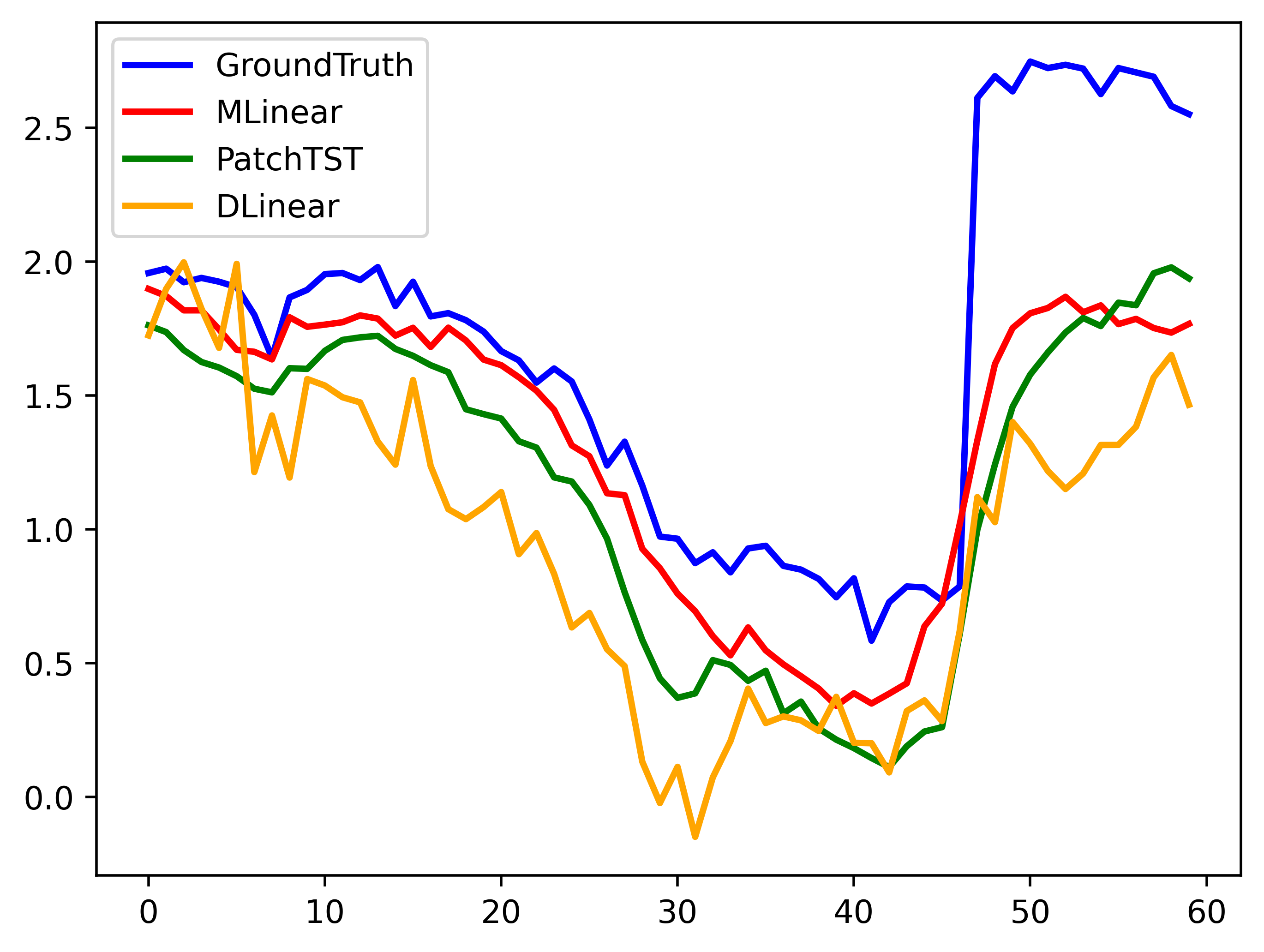}}
	\subfloat[]{\includegraphics[width=0.33\textwidth]{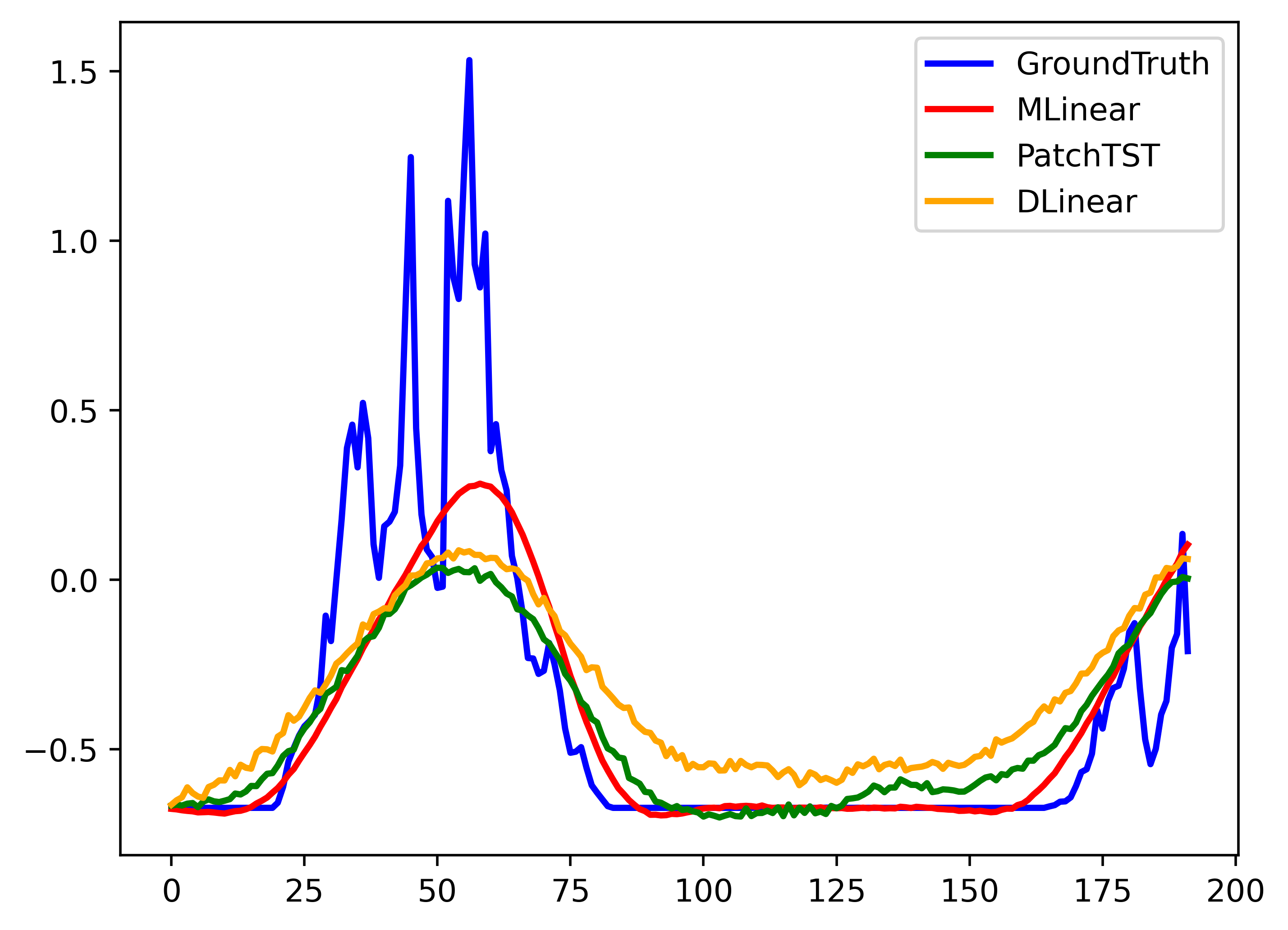}}
	\caption{Visualization Analysis: From left to right, the ETTm1 dataset, the ILI dataset, and the weather dataset are displayed. The forecasting horizons for these three datasets are 192, 60, and 192, respectively.}
	\label{}
\end{figure*}
\subsubsection{Deeply supervised ablation experiment}
We compare the effectiveness of deep supervision on four different loss functions to fully validate its generality and scalability across different loss functions. 

From Table~\ref{DEEPSUP}, it can be seen that in most cases, deep supervision achieves a general performance improvement compared to simple single-loss methods. The only exceptions are in the case of MAE on ETTh1 with a prediction length of 720, and on ETTm2 with a prediction length of 192, as well as on M-Loss on ETTm2 with a prediction length of 92, where no significant performance improvement is observed. It is noteworthy that deep supervision does not introduce any additional parameters nor does it affect inference efficiency, making this general improvement highly significant.
\subsection{Different fusion methods for mixing CI and CD}
We deeply explore different aggregation methods and conduct three sets of comparisons to achieve two research objectives: (1) Is attention aggregation more effective than linear aggregation? (2) Can stacked attention layers aggregation improve prediction performance? All three experiments use the full version of MLinear, but replace the aggregation method with Efficient self-attention (ATT), linear, and 3-layer ATT (3ATT).

From Table~\ref{3AGG}, it can be seen that on the four datasets ETTh1, ETTh2, ETTm1, and ETTm2, 1ATT achieves the best performance 19 times, while simple aggregation only achieves the best performance 11 times. It is noteworthy that although the number of ATT layers is increased, the performance of 3-layer ATT is not better than that of 1ATT, and even the number of best performances is lower. This suggests that stacking too many parameters may reduce the generalization ability of MLinear on LSTF datasets with high information density.
\subsection{Look-back window analysis}
\begin{figure*}[htbp]
	\centering
	\includegraphics[width=1\textwidth]{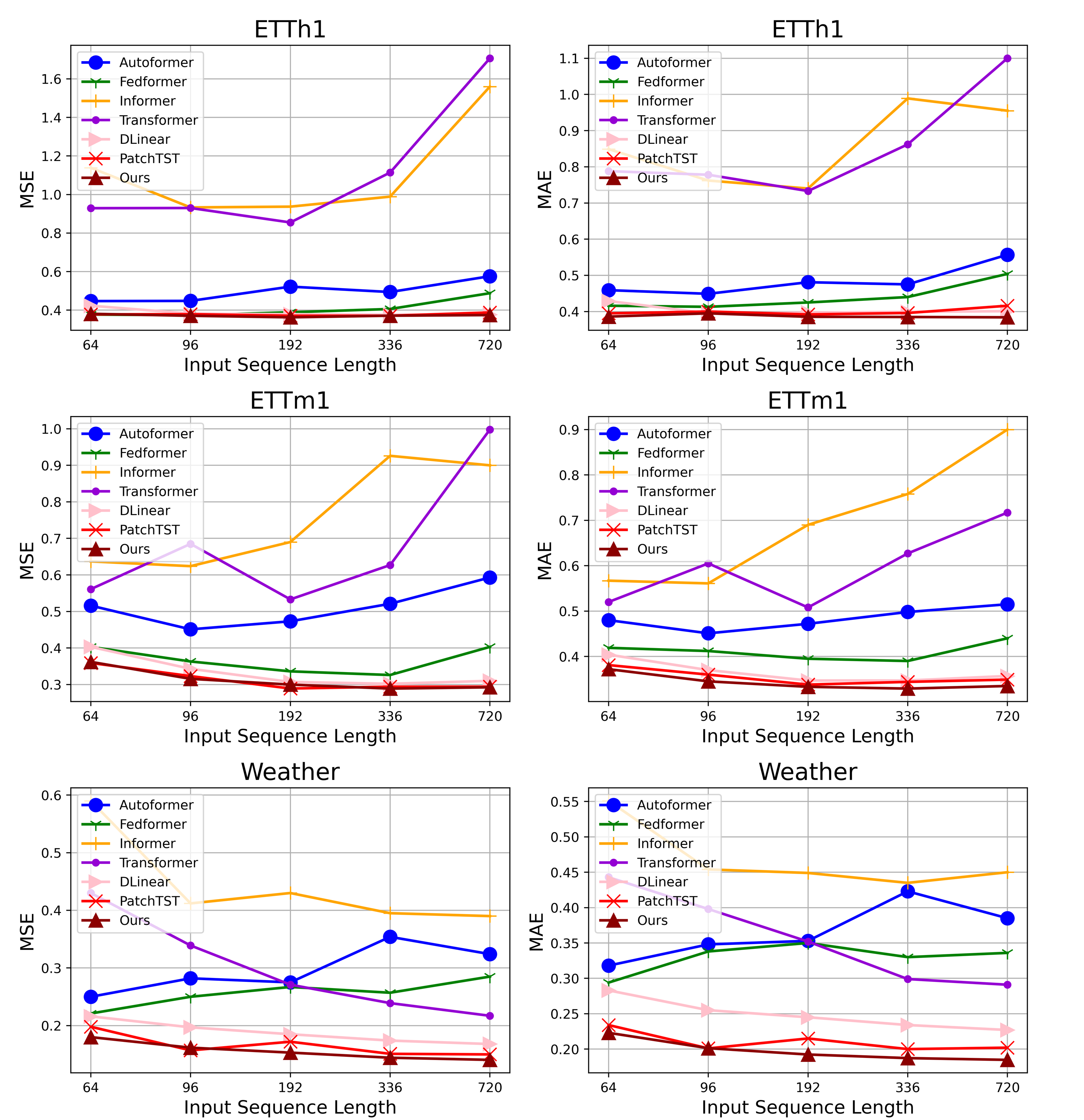}
	\caption{Forecasting performance (MSE and MAE) with varying look-back windows on Weather. The look-back windows are selected to be 64, 96, 192, 336, 720, and the prediction horizons are 96.}
	\label{lookall}
\end{figure*}
Intuitively, analyzing longer sequences from the past can increase our predictive knowledge. However, previous works PatchTST~\cite{nie2022time} and DLinear~\cite{zeng2022transformers} have indicated a significant performance drop for Transformer-based models when handling longer inputs. To investigate this, we conducted experiments on three datasets using MSE and MAE metrics, where we varied the look-back window while keeping the prediction horizon fixed at 96 time steps. Our results are shown in Figure~\ref{lookall}. 
Two trends are revealed in Figure~\ref{lookall}: (1) Transformer, Informer, Fedformer, and Autoformer all suffer significant performance drops as the input sequence length increases to 336 and 720. (2) DLinear, PatchTST, and MLinear (our proposed model) show improved predictive performance as the input sequence length increases, implying that these models can benefit from longer input sequences in terms of knowledge capacity. Additionally, MLinear performs relatively better than PatchTST and DLinear in absolute performance.
\subsection{Efficiency Comparison}
We conducted efficiency comparisons between different models.
In order to make the comparison results more distinct, all tests were conducted using the code published by their respective authors on a RTX 3070 GPU the input sequence length is 336 and the forecasting length is 96 in ETTh1.

Table~\ref{eff} shows that both DLinear and MLinear (our proposed model) achieve significant efficiency gains compared to other baseline methods. In terms of parameter count, MLinear has a smaller parameter count than DLinear. However, the introduction of attention mechanisms for tuning the CI and CD increases computational demand in both training and inference time, which results in no significant difference in training and inference time compared to DLinear.
\begin{table*}[]
	\centering
	\caption{Efficiency Comparison (Per epoch).}
	\label{eff}
	
	\begin{tabular}{cccc}
		\hline
		& Parameter & Training Time& Inference Time \\ \hline
		Informer      & 11.32M    & 22.97s     &2.87s    \\ \hline
		Autoformer    & 10.53M   & 32.84s    &7.27s    \\ \hline
		Transformer   & 10.54M    & 26.83s    &3.13s    \\ \hline
		PatchTST      & 20.78M    & 21.76s  & 2.57s   \\\hline
		DLinear~(Individual)       & 0.452M    & 1.03s    &0.29s    \\\hline
		MLinear~(Ours) & 0.277M    & 1.14s    &0.29s   \\\hline
	\end{tabular}
\end{table*}


\section{Conclusion and future work}
In this paper, we propose MLinear and address an unresolved question: Can we effectively utilize channel-independence (CI) and channel-dependence (CD) to improve the performance of long sequence time series forecasting (LSTF)? The two design principles of MLinear provide a definitive answer to this question: 
(1) Time-semantics-aware tuning CI and CD according to input semantics; (2) Treating CI and CD as separate predictors and applying deep supervision to them. Additionally, we provide an effective loss function for time series forecasting tasks through experimentation. Experimental results on seven widely used datasets demonstrate that our proposed MLinear achieves significant predictive advantages over state-of-the-art baseline methods. Furthermore, MLinear's efficiency is also top-tier, significantly outperforming current Transformer-based methods in both parameter usage and inference time.

{\small
\bibliographystyle{splncs04}
\bibliography{meng}
}

\end{document}